\documentclass{article}

\PassOptionsToPackage{numbers, compress}{natbib}
\usepackage[final]{neurips_2022}

\usepackage[utf8]{inputenc} 
\usepackage[T1]{fontenc}    
\usepackage{hyperref}       
\usepackage{url}            
\usepackage{booktabs}       
\usepackage{amsfonts}       
\usepackage{nicefrac}       
\usepackage{microtype}      
\usepackage{xcolor}         
\usepackage{graphicx}

\usepackage{multirow}
\usepackage{epsfig}
\usepackage{amsmath}
\usepackage{amssymb}
\usepackage{subfigure}
\usepackage{marvosym}
\usepackage{color}
\usepackage{threeparttable}
\usepackage{algorithm}
\usepackage{algpseudocode}
\usepackage{setspace}
\usepackage{amsthm}

\theoremstyle{definition}
\newtheorem*{theorem}{Theorem}
\newtheorem{assumption}{Assumption}
\newtheorem*{definition}{Definition}

\title{Non-stationary Transformers:\\ Exploring the Stationarity in Time Series Forecasting}

\author{
  Yong Liu\thanks{Equal Contribution},
  Haixu Wu\footnotemark[1],
  Jianmin Wang,
  Mingsheng Long\textsuperscript{\Letter} \\
  School of Software, BNRist, Tsinghua University, China \\
  {\small \texttt{\{liuyong21,whx20\}@mails.tsinghua.edu.cn},  \texttt{\{jimwang,mingsheng\}@tsinghua.edu.cn}}
}

\begin{document}

\maketitle

\begin{abstract}
Transformers have shown great power in time series forecasting due to their global-range modeling ability. However, their performance can degenerate terribly on non-stationary real-world data in which the joint distribution changes over time. Previous studies primarily adopt stationarization to attenuate the non-stationarity of original series for better predictability. But the stationarized series deprived of inherent non-stationarity can be less instructive for real-world bursty events forecasting. This problem, termed \emph{over-stationarization} in this paper, leads Transformers to generate indistinguishable temporal attentions for different series and impedes the predictive capability of deep models. To tackle the dilemma between series predictability and model capability, we propose \emph{Non-stationary Transformers} as a generic framework with two interdependent modules: Series Stationarization and De-stationary Attention. Concretely, Series Stationarization unifies the statistics of each input and converts the output with restored statistics for better predictability. To address the over-stationarization problem, De-stationary Attention is devised to recover the intrinsic non-stationary information into temporal dependencies by approximating distinguishable attentions learned from raw series. Our Non-stationary Transformers framework consistently boosts mainstream Transformers by a large margin, which reduces MSE by 49.43\% on Transformer, 47.34\% on Informer, and 46.89\% on Reformer, making them the state-of-the-art in time series forecasting. Code is available at this repository: \url{https://github.com/thuml/Nonstationary_Transformers}.
\end{abstract}

\section{Introduction} \label{sec:intro}
Time series forecasting has become increasingly ubiquitous in real-world applications, such as weather forecasting, energy consumption planning, and financial risk assessment. Recently, Transformers~\cite{NIPS2017_3f5ee243} have achieved progressive breakthrough on extensive areas~\cite{Devlin2019BERTPO,dosovitskiy2021an,chen2021decisiontransformer,liu2021Swin}. Especially in time series forecasting, credited to their stacked structure and the capability of attention mechanisms, Transformers can naturally capture the temporal dependencies from deep multi-level features~\cite{haoyietal-informer-2021,kitaev2020reformer,2019Enhancing,wu2021autoformer}, thereby fitting the series forecasting task perfectly.

Despite the remarkable architectural design, it is still challenging for Transformers to predict real-world time series because of the non-stationarity of data. Non-stationary time series is characterized by the continuous change of statistical properties and joint distribution over time, which makes the time series less predictable~\cite{Anderson1976TimeSeries2E,hyndman2018forecasting}. Besides, it is a fundamental problem to make deep models generalize well on a varying  distribution~\cite{pan2009survey, li2017deeper, ahuja2021invariance}. In previous work, it is generally acknowledged to pre-process the time series by stationarization~\cite{adanorm, passalis2019deep, kim2022reversible}, which can attenuate the non-stationarity of raw time series for better predictability and provide more stable data distribution for deep models. 


However, non-stationarity is the inherent property of real-world time series and also good guidance for discovering temporal dependencies for forecasting. Experimentally, we observe that training on the stationarized series will undermine the distinction of attentions learned by Transformers. While vanilla Transformers~\cite{NIPS2017_3f5ee243} can capture distinct temporal dependencies from different series in Figure~\ref{fig:motivation}(a), Transformers trained on the stationarized series tend to generate indistinguishable attentions in Figure~\ref{fig:motivation}(b). This problem, named by the \emph{over-stationarization}, will bring unexpected side-effect that makes Transformers fail to capture eventful temporal dependencies, limit the model's predictive ability, and even induce the model to generate outputs with huge non-stationarity deviation from the ground truth. Thus, \emph{how to attenuate time series non-stationarity towards better predictability and mitigate the over-stationarization problem for model capability simultaneously} is the key problem to further improve the performance of forecasting.

In this paper, we explore the effect of stationarization in time series forecasting and propose \emph{Non-stationary Transformers} as a general framework, which empowers Transformer~\cite{NIPS2017_3f5ee243} and its efficient variants~\cite{kitaev2020reformer,haoyietal-informer-2021,wu2021autoformer} with great predictive ability for real-world time series. The proposed framework involves two interdependent modules: Series Stationarization to increase the predictability of non-stationary series and De-stationary Attention to alleviate over-stationarization. Technically, Series Stationarization adopts a simple but effective normalization strategy to unify the key statistics of each series without extra parameters. And De-stationary Attention approximates the attention of unstationarized data and compensates the intrinsic non-stationarity of raw series. Benefiting from the above designs, Non-stationary Transformers can take advantage of the great predictability of stationarized series and crucial temporal dependencies discovered from original non-stationary data. Our method achieves state-of-the-art performance on six real-world benchmarks and can generalize to various Transformers for further improvement. The contributions lie in three folds:

\begin{figure}[tbp]
  \includegraphics[width=.98\columnwidth]{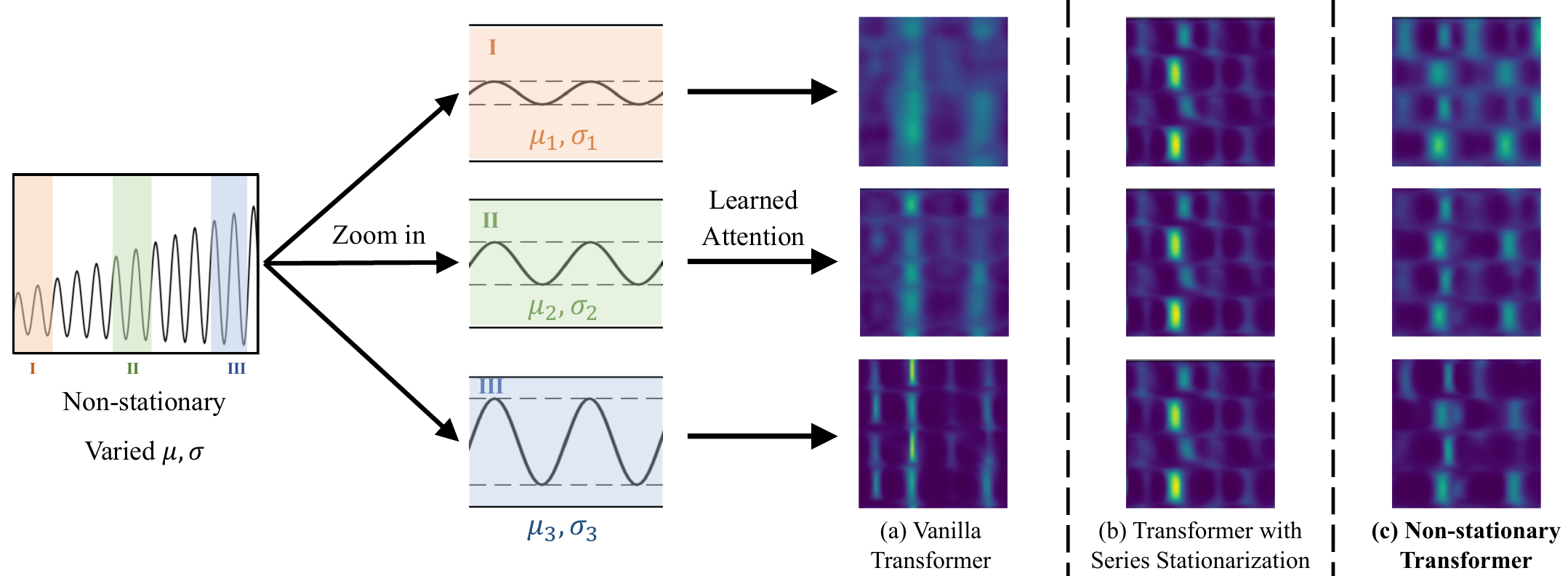}
  \centering
  \vspace{-5pt}
  \caption{{Visualization of learned temporal attentions} for different series with varied mean $\mu$ and standard deviation $\sigma$.
  (a) is from the vanilla Transformer~\cite{NIPS2017_3f5ee243} trained on raw series. (b) is from the Transformer trained on stationarized series, which presents similar attentions. (c) is from Non-stationary Transformers, which involves De-stationary Attention to avoid over-stationarization.}
  \label{fig:motivation}
  \vspace{-10pt}
\end{figure}

\begin{itemize}
    \item We refine that the predictive capability of non-stationary series is essential in real-world forecasting. By detailed analysis, we find out that current stationarization approaches will lead to the over-stationarization problem, limiting the predictive capability of Transformers.
    \item We propose Non-stationary Transformers as a generic framework, including Series Stationarization to make the series more predictable and De-stationary Attention to avoid the over-stationarization problem by re-incorporating the non-stationarity of original series.
    \item Non-stationary Transformers consistently boosts four mainstream Transformers by a large margin and achieves state-of-the-art performance on six real-world benchmarks.
\end{itemize}

\section{Related Work}

\subsection{Deep Models for Time Series Forecasting}
In recent years, deep models with elaboratively designed architectures have achieved great progress in time series forecasting. RNN-based models~\cite{Wen2017AMQ, 2017Long, Maddix2018DeepFW, Rangapuram2018DeepSS, Flunkert2017DeepARPF} are proposed for application in an autoregressive manner for sequence modeling, but the recurrent structure can suffer from modeling long-term dependency. Soon afterward, Transformer~\cite{NIPS2017_3f5ee243} emerges and shows great power in sequence modeling. To overcome the quadratic computation growth on sequence length, subsequent works aim to reduce Self-Attention's complexity. Especially in time series forecasting, Informer~\cite{haoyietal-informer-2021} extends Self-Attention with KL-divergence criterion to select dominant queries. Reformer~\cite{kitaev2020reformer} introduces local-sensitive hashing (LSH) to approximate attention by allocated similar queries. Not only improved by reduced complexity, the following models further develop delicate building blocks for time series forecasting. Autoformer~\cite{wu2021autoformer} fuses the decomposition blocks into a canonical structure and develops Auto-Correlation to discover series-wise connections. Pyraformer~\cite{liu2021pyraformer} designs pyramid attention module (PAM) to capture temporal dependencies with different hierarchies. Other deep but Transformer-free models also achieve remarkable performance. N-BEATS~\cite{oreshkin2019n} proposes the explicit decomposition of trend and seasonal terms with strong interpretability. N-HiTS~\cite{challu2022n} introduces hierarchical layout and multi-rate sampling for tackling time series with respective frequency bands. In this paper, different from previous works focusing on architectural design, we analyze the series forecasting task from the basic view of stationarity, which is an essential property of time series~\cite{Anderson1976TimeSeries2E,hyndman2018forecasting}. It is also notable that as a general framework, our proposed Non-stationary Transformers can be easily applied to various Transformer-based models.

\subsection{Stationarization for Time Series Forecasting}
While stationarity is important to the predictability of time series~\cite{Anderson1976TimeSeries2E,hyndman2018forecasting}, real-world series always present non-stationarity. To tackle this problem, the classical statistical method ARIMA~\cite{Box1970TimeSA, Box1968SomeRA} stationarizes the time series through differencing. As for deep models,
since the distribution-varying problem accompanied by non-stationarity makes deep forecasting even more intractable, stationarization methods are widely explored and always adopted as the pre-processing for deep model inputs. Adaptive Norm~\cite{adanorm} applies z-score normalization for each series fragment by global statistics of a sampled set. DAIN~\cite{passalis2019deep} employs a nonlinear neural network to adaptively stationarize time series with observed training distribution. RevIN~\cite{kim2022reversible} introduces a two-stage instance normalization~\cite{ulyanov2016instance} that transforms model input and output respectively to reduce the discrepancy of each series. In contrast, we find out that directly stationarizing time series will damage the model's capability of modeling specific temporal dependency. Therefore, unlike previous methods, in addition to the stationarization, Non-stationary Transformers further develops De-stationary Attention to bring the intrinsic non-stationarity of the raw series back to attention.

\section{Non-stationary Transformers}

As aforementioned, stationarity is an important element of time series predictability. Previous ``direct stationarization'' designs can 
attenuate non-stationarity of series for better predictability, but they obviously neglect inherent properties of real-world series, which will result in the over-stationarization problem as stated in Figure~\ref{fig:motivation}. To deal with the dilemma, we go beyond previous works and propose \emph{Non-stationary Transformers} as a generic framework.
Our model involves two complementary parts: Series Stationarization to attenuate time series non-stationarity and De-stationary Attention to re-incorporate non-stationary information of raw series.
Empowered by these designs, Non-stationary Transformers can improve data predictability and maintain model capability simultaneously.
 
\subsection{Series Stationarization}\label{sec:sm}
Non-stationary time series make the forecasting task intractable for deep models because it is hard for them to generalize well on series with changed statistics during inference, typically varied mean and standard deviation.
The pilot work, RevIN~\cite{kim2022reversible} applies instance normalization with learnable affine parameters to each input and restores the statistics to the corresponding output, which makes each series follow a similar distribution. Experimentally, we find that this design also works well without learnable parameters. Thus, we propose a more straightforward but effective design to wrap Transformers as the base model without extra parameters, naming by Series Stationarization. As is shown in Figure~\ref{fig:arch}, it contains two corresponding operations: Normalization module at first to deal with the non-stationary series caused by varied mean and standard deviation, and De-normalization module at the end to transform the model outputs back with original statistics. Here are the details.

\vspace{-5pt}
\paragraph{Normalization module} To attenuate the non-stationarity of each input series, we conduct normalization on the temporal dimension by a sliding window over time. For each input series $\mathbf{x}=[x_1,x_2,...,x_S]^{\top} \in\mathbb{R}^{S \times C}$, we transform it by translation and scaling operations and obtain $\mathbf{x^\prime}=[x^\prime_1,x^\prime_2,...,x^\prime_S]^{\top} \in\mathbb{R}^{S \times C}$, where $S$ and $C$ denote the sequence length and variable number respectively. The Normalization module can be formulated as follows:
\begin{equation}
    \mu_{\mathbf{x}} = \frac{1}{S}\sum_{i=1}^S x_i,\ \sigma_{\mathbf{x}}^2 =\frac{1}{S} \sum_{i=1}^S (x_i-\mu_{\mathbf{x}})^2,\ x^\prime_i = \frac{1}{\sigma_{\mathbf{x}}} \odot (x_i-\mu_{\mathbf{x}}),
\end{equation}
where $\mu_{\mathbf{x}}, \sigma_{\mathbf{x}} \in\mathbb{R}^{C \times 1} $, $\frac{1}{\sigma_{\mathbf{x}}}$ means the element-wise division and $\odot$ is the element-wise product. Note that Normalization module decreases the distributional discrepancy among each input time series, making the distribution of the model input more stable.

\vspace{-5pt}
\paragraph{De-normalization module} As shown in Figure~\ref{fig:arch}, after the base model $\mathcal{H}$ predicting the future value with length-$O$, we adopt De-normalization to transform the model output $\mathbf{y^\prime}=[y^\prime_1, y^\prime_2, ..., y^\prime_O]^{\top} \in\mathbb{R}^{O \times C}$ with $\sigma_{\mathbf{x}}$ and $\mu_{\mathbf{x}}$ and obtain $\hat{\mathbf{y}}=[\hat{y}_1, \hat{y}_2, ..., \hat{y}_O]^{\top}$ as the eventual forecasting results. The De-normalization module can be formulated as follows:
\begin{equation}
    \mathbf{y^\prime} = \mathcal{H}(\mathbf{x^\prime}),\ \hat{y}_i = \sigma_{\mathbf{x}} \odot y^\prime_i+\mu_{\mathbf{x}}.
\end{equation}
By means of the two-stage transformation, the base models will receive stationarized inputs, which follow a stable distribution and are easier to generalize. This design also makes the model equivariant to translational and scaling perturbance of time series, thereby benefiting real-world series forecasting.

\vspace{-5pt}
\subsection{De-stationary Attention}\label{sec:da}
While the statistics of each time series are explicitly restored to the corresponding prediction, the non-stationarity of the original series cannot be fully recovered only by De-normalization. For instance, Series Stationarization can generate the same stationarized input $\mathbf{x}^{\prime}$ from distinct time series $\mathbf{x}_1$, $\mathbf{x}_2$ (i.e. $\mathbf{x}_2=\alpha \mathbf{x}_1 + \beta$), and the base model will get identical attention that fails to capture crucial temporal dependencies entangled with non-stationarity (Figure~\ref{fig:motivation}). In other words, the undermined effects caused by over-stationarization happen inside the deep model, especially in the calculation of attention. Furthermore, non-stationary time series are fragmented and normalized into several series chunks with the same mean and variance, which follow more similar distributions than the raw data before stationarization. Thus, the model is more likely to generate over-stationary and uneventful outputs, which is irreconcilable with the natural non-stationarity of the original series.

To tackle the over-stationarization problem caused by Series Stationarization, we propose a novel De-stationary Attention mechanism, which can approximate the attention that is obtained without stationarization and discover the particular temporal dependencies from original non-stationary data.
\begin{figure}[tbp]
  \includegraphics[width=.9\columnwidth]{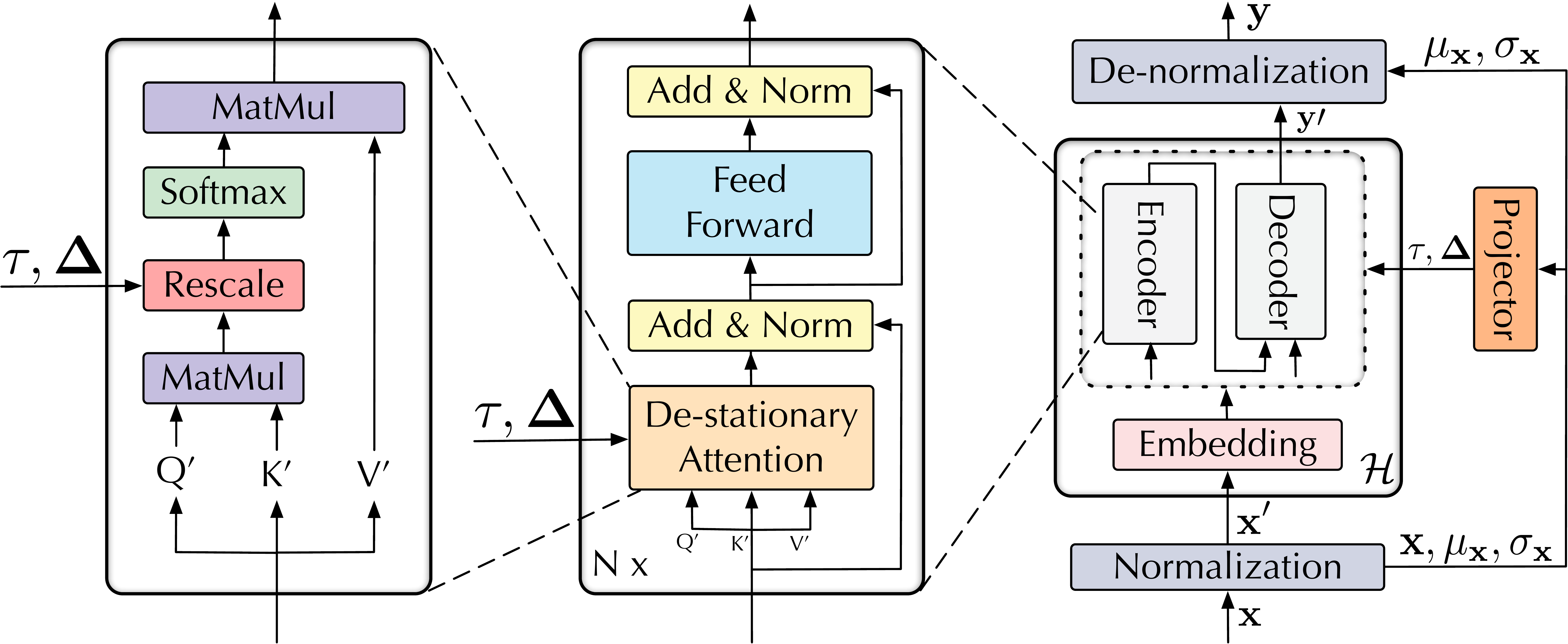}
  \centering
  \vspace{-5pt}
  \caption{Non-stationary Transformers. Series Stationarization is adopted as a wrapper on the base model to normalize each incoming series and de-normalize the output. De-stationary Attention replaces the original Attention mechanism to approximate attention learned from unstationarized series, which rescales current temporal dependency weights with learned de-stationary factors $\tau,\mathbf{\Delta}$.}
  \label{fig:arch}
  \vspace{-5pt}
\end{figure}

\paragraph{Analysis of the plain model} As mentioned above, the over-stationarization problem is caused by the vanishment of inherent non-stationarity information, which will make the base model fail to capture eventful temporal dependencies for forecasting. Therefore, we try to approximate the attention learned from the original non-stationary series. We start from the formula of Self-Attention~\cite{NIPS2017_3f5ee243}:
\begin{equation}
\label{equ:attn}
    \operatorname{Attn}(\mathbf{Q}, \mathbf{K}, \mathbf{V} )=\operatorname{Softmax}\left(\frac{\mathbf{Q} \mathbf{K}^\top}{\sqrt{d_{k}}}\right) \mathbf{V},
\end{equation}
where $\mathbf{Q},\mathbf{K},\mathbf{V}\in\mathbb{R}^{S\times d_k}$ are length-$S$ queries, keys and values of $d_k$-dimension respectively, and $\operatorname{Softmax}(\cdot)$ is conducted row by row. To simplify the analysis, we assume the embedding and feed-forward layers $f$ to hold the linear properties\footnote{Function $f$ has the linear property if it satisfies that $f(ax+by)=af(x)+bf(y)$, where $a, b$ are scalar constants and $x,y$ are vector variables.} and $f$ is conducted separately on each time point, that is, each query token in $\mathbf{Q} = [q_1, q_2, ..., q_S]^\top$ can be calculated as $q_i=f(x_i)$ with respect to the input series $\mathbf{x}=[x_1,x_2,\cdots,x_S]^\top$. Since it is a convention to conduct normalization on each time series variable to avoid certain variable that dominates the scale, we can further assume each variable of series $\mathbf{x}$ shares the same variance, and thus original $\sigma_{\mathbf{x}}\in\mathbb{R}^{C\times 1}$ is reduced to a scalar. After Normalization module, the model receives the stationarized input $\mathbf{x}^{\prime}=(\mathbf{x}- \mathbf{1} \mu_{\mathbf{x}}^{\top} )/\sigma_{\mathbf{x}}$, where $\mathbf{1} \in \mathbb{R}^{S\times 1}$ is an all-ones vector. Based on the linear property assumption, it can be proved that the Attention layer will receive $\mathbf{Q}^{\prime}= [f(x_1^\prime), ..., f(x_S^\prime)]^\top =(\mathbf{Q}- \mathbf{1}\mu_{\mathbf{Q}}^\top)/\sigma_{\mathbf{x}}$, where $\mu_{\mathbf{Q}} \in\mathbb{R}^{d_k \times 1}$ is the mean of $\mathbf{Q}$ along the temporal dimension (See Appendix~\ref{app:proof} for a detailed proof). And so is the corresponding transformed $\mathbf{K}^{\prime}, \mathbf{V}^{\prime}$. Without Series Stationarization, the input of $\operatorname{Softmax}(\cdot)$ in Self-Attention should be $\mathbf{Q}\mathbf{K}^{\top}/\sqrt{d_k}$, while now the attention is calculated based on $\mathbf{Q}^{\prime}, \mathbf{K}^{\prime}$:
\begin{equation}
\label{equ:insight0}
\begin{split}
    \mathbf{Q}^{\prime}\mathbf{K}^{\prime\top} &=\frac{1}{\sigma_{\mathbf{x}}^2} \left( \mathbf{Q}\mathbf{K}^{\top}-
    \mathbf{1} (\mu_{\mathbf{Q}}^{\top} \mathbf{K}^{\top}) -
    (\mathbf{Q} \mu_{\mathbf{K}}) \mathbf{1}^{\top} + 
    \mathbf{1} (\mu_{\mathbf{Q}}^{\top} \mu_{\mathbf{K}}) \mathbf{1}^{\top} \right), \\
    \operatorname{Softmax}\left( \frac{\mathbf{Q} \mathbf{K}^{\top}}{\sqrt{d_k}}\right) &= 
    \operatorname{Softmax}\left( \frac{
    \sigma_{\mathbf{x}}^2 \  \mathbf{Q^\prime}\mathbf{K^\prime}^{\top} + 
    \mathbf{1} (\mu_{\mathbf{Q}}^{\top} \mathbf{K}^{\top}) +
    (\mathbf{Q} \mu_{\mathbf{K}}) \mathbf{1}^{\top} -
    \mathbf{1} (\mu_{\mathbf{Q}}^{\top} \mu_{\mathbf{K}}) \mathbf{1}^{\top}}{\sqrt{d_k}}
    \right).
\end{split}
\end{equation}
 We find that $\mathbf{Q} \mu_{\mathbf{K}} \in \mathbb{R}^{S\times 1}$ and $\mu_{\mathbf{Q}}^{\top} \mu_{\mathbf{K}}\in \mathbb{R}$, and they are repeatedly operated on each column and element of $\sigma_{\mathbf{x}}^2\mathbf{Q^\prime}\mathbf{K^\prime}^{\top}\in \mathbb{R}^{S\times S}$ respectively. Since $\operatorname{Softmax}(\cdot)$ is invariant to the same translation on the row dimension of input, we have the following equation: 
\begin{equation}\label{equ:insight}
     \operatorname{Softmax}\left( \frac{\mathbf{Q} \mathbf{K}^{\top}}{\sqrt{d_k}}\right)  = \operatorname{Softmax}\left( \frac{ \sigma_{\mathbf{x}}^2 \  \mathbf{Q^\prime}\mathbf{K^\prime}^{\top} + \mathbf{1} \mu_{\mathbf{Q}}^{\top} \mathbf{K}^{\top}}{\sqrt{d_k}}\right).
\end{equation}
Equation~\ref{equ:insight} deduces a direct expression of the attention  $\operatorname{Softmax}\left( \mathbf{Q} \mathbf{K}^{\top} / \sqrt{d_k} \right)$ learned from raw series $\mathbf{x}$. Except for the current $\mathbf{Q}^\prime,\mathbf{K}^\prime$ from stationarized series $\mathbf{x^{\prime}}$, this expression also requires the non-stationary information $\sigma_{\mathbf{x}},\mu_{\mathbf{Q}},\mathbf{K}$ that are eliminated by Series Stationarization.

\paragraph{De-stationary Attention} To recover the original attention on non-stationary series, we attempt to bring the vanished non-stationary information back to its calculation. Based on Equation~\ref{equ:insight}, the key is to approximate the positive scaling scalar $\tau=\sigma_{\mathbf{x}}^2\in\mathbb{R}^{+}$ and shifting vector $\mathbf{\Delta}=\mathbf{K}\mu_\mathbf{Q} \in\mathbb{R}^{S\times 1}$, which are defined as \emph{de-stationary factors}.
Since the strict linear property hardly holds for a deep model, other than estimating and utilizing real factors with great effort, we try to learn de-stationary factors directly from the statistics of unstationarized $\mathbf{x}, \mathbf{Q}$ and $\mathbf{K}$ by a simple but effective multi-layer perceptron layer. As we can only discover limited non-stationary information from current $\mathbf{Q}^{\prime}, \mathbf{K}^{\prime}$, the unique and reasonable source to compensate non-stationarity is the original $\mathbf{x}$ without being normalized. Thus, as a direct deep learning implementation of Equation~\ref{equ:insight}, we apply a multi-layer perceptron as the projector to learn de-stationary factors $\tau, \mathbf{\Delta}$ from the statistics $\mu_{\mathbf{x}}, \sigma_{\mathbf{x}}$ of unstationarized $\mathbf{x}$ individually. And the De-stationary Attention is calculated as follows: 
\begin{equation}\label{equ:nsattn}
  \begin{split}
    & \log{\tau} = \operatorname{MLP}(\sigma_\mathbf{x}, \mathbf{x}), \mathbf{\Delta} = \operatorname{MLP}(\mu_\mathbf{x}, \mathbf{x}), \\
    \operatorname{Attn} (\mathbf{Q^\prime}, & \mathbf{K^\prime}, \mathbf{V^\prime}, \tau, \mathbf{\Delta} ) = \operatorname{Softmax}\left(\frac{\tau \ \mathbf{Q^\prime} \mathbf{K^\prime}^\top+ \mathbf{1}\mathbf{\Delta}^{\top}}{\sqrt{d_{k}}}\right) \mathbf{V^\prime},
  \end{split}
\end{equation}
where the de-stationary factors $\tau$ and $\mathbf{\Delta}$ are shared by De-stationary Attention of all layers  (Figure~\ref{fig:arch}). De-stationary Attention mechanism learns the temporal dependencies from both stationarized series $\mathbf{Q}'$, $\mathbf{K'}$ and non-stationary series $\mathbf{x}$, $\mu_{\mathbf{x}}$, $\sigma_{\mathbf{x}}$, and multiplies by the stationarized values $\mathbf{V^\prime}$. Therefore, it can benefit from the predictability of stationarized series and maintain the inherent temporal dependencies of raw series simultaneously.

\paragraph{Overall architecture} Following the prior use of Transformers~\cite{haoyietal-informer-2021,wu2021autoformer} in time series forecasting, we adopt the standard Encoder-Decoder structure (Figure \ref{fig:arch}), where the encoder is to extract information from past observations, and the decoder is to aggregate past information and refine the prediction from simple initialization. The canonical Non-stationary Transformer is wrapped by Series Stationarization to both the input and output of vanilla Transformer~\cite{NIPS2017_3f5ee243}, and replacing the Self-Attention by our proposed De-stationary Attention, which can boost the non-stationary series predictive capability of the base model. For the Transformer variants~\cite{kitaev2020reformer,haoyietal-informer-2021,wu2021autoformer}, we transform the terms inside $\operatorname{Softmax}(\cdot)$ with the de-stationary factors $\tau$, $\mathbf{\Delta}$ to re-integrate the non-stationary information (See Appendix~\ref{app:impl} for the implementation details).

\vspace{-5pt}
\section{Experiments} \label{sec:exp}
\vspace{-5pt}
We conduct extensive experiments to evaluate the performance of Non-stationary Transformers on six real-world time series forecasting benchmarks and further validate the generality of the proposed framework on various mainstream Transformer variants.
\vspace{-5pt}
\paragraph{Datasets} Here are the descriptions of the datasets: 
(1) \textbf{Electricity}~\cite{ecldata} records the hourly electricity consumption of 321 clients from 2012 to 2014.
(2) \textbf{ETT}~\cite{haoyietal-informer-2021} contains the time series of oil temperature and power load collected by electricity transformers from July 2016 to July 2018. ETTm1 /ETTm2 are recorded every 15 minutes, and ETTh1/ETTh2 are recorded every hour.
(3) \textbf{Exchange}~\cite{2018Modeling} collects the panel data of daily exchange rates from 8 countries from 1990 to 2016. 
(4) \textbf{ILI}~\cite{ilidata} collects the ratio of influenza-like illness patients versus the total patients in one week, which is reported weekly by Centers for Disease Control and Prevention of the United States from 2002 and 2021.
(5) \textbf{Traffic}~\cite{trafficdata} contains hourly road occupancy rates measured by 862 sensors on San Francisco Bay area freeways
from January 2015 to December 2016.
(6) \textbf{Weather}~\cite{weatherdata} includes meteorological time series with 21 weather indicators collected every 10 minutes from the Weather Station of the Max Planck Biogeochemistry Institute in 2020.

Especially, in this paper, we adopt the Augmented Dick-Fuller (ADF) test statistic~\cite{adftest} as the metric to quantitatively measure the \emph{degree of stationarity}. A smaller ADF test statistic indicates a higher degree of stationarity, which means the distribution is more stable. Table~\ref{tab:Datasets} summarizes the overall statistics of the datasets and lists them in ascending order by degree of stationarity. We follow the standard protocol that divides each dataset into the training, validation, and testing subsets according to the chronological order. The split ratio is 6:2:2 for the ETT dataset and 7:1:2 for others.

\begin{table}[htbp]
  \caption{Summary of datasets. Smaller ADF test statistic indicates more stationary dataset.}\label{tab:Datasets}
  \vspace{-5pt}
  \centering
  \begin{small}
    \begin{threeparttable}
    \setlength{\tabcolsep}{8.5pt}
    \begin{tabular}{lcccc}
    \toprule
    \multicolumn{1}{l}{Dataset} & \multicolumn{1}{l}{Variable Number} & \multicolumn{1}{l}{Sampling Frequency} & \multicolumn{1}{l}{Total Observations} & \multicolumn{1}{l}{ADF Test Statistic} \\ \toprule
    Exchange                    & 8                                      & \multicolumn{1}{c}{1 Day}              & \multicolumn{1}{c}{7,588}             & \multicolumn{1}{c}{-1.889}             \\
    ILI                     & 7                         & \multicolumn{1}{c}{1 Week}             & \multicolumn{1}{c}{966}               & \multicolumn{1}{c}{-5.406}             \\
    ETTm2                       & 7                                      & \multicolumn{1}{c}{15 Minutes}         & \multicolumn{1}{c}{69,680}            & \multicolumn{1}{c}{-6.225}            \\
    Electricity                 & 321                                    & \multicolumn{1}{c}{1 Hour}             & \multicolumn{1}{c}{26,304}            & \multicolumn{1}{c}{-8.483}             \\
    Traffic                     & 862                                    & \multicolumn{1}{c}{1 Hour}             & \multicolumn{1}{c}{17,544}            & \multicolumn{1}{c}{-15.046}            \\
    Weather                     & 21                                     & \multicolumn{1}{c}{10 Minutes}         & \multicolumn{1}{c}{52,695}            & \multicolumn{1}{c}{-26.661}            \\ \bottomrule
    \end{tabular}
  \end{threeparttable}
  \end{small}
  \vspace{-5pt}
\end{table}

\paragraph{Baselines} We evaluate the vanilla Transformer~\cite{NIPS2017_3f5ee243} equipped by the Non-stationary Transformers framework in both multivariate and univariate settings to demonstrate its effectiveness. For multivariate forecasting, we include six state-of-the-art deep forecasting models: Autoformer~\cite{wu2021autoformer}, Pyraformer~\cite{liu2021pyraformer}, Informer \cite{haoyietal-informer-2021}, LogTrans~\cite{2019Enhancing}, Reformer~\cite{kitaev2020reformer} and LSTNet~\cite{2018Modeling}. For univariate forecasting, we include seven competitive baselines: N-HiTS~\cite{challu2022n}, N-BEATS~\cite{oreshkin2019n}, Autoformer~\cite{wu2021autoformer}, Pyraformer~\cite{liu2021pyraformer}, Informer~\cite{haoyietal-informer-2021}, Reformer~\cite{kitaev2020reformer} and ARIMA~\cite{Box1970TimeSA}. In addition, we adopt the proposed framework on both the canonical and efficient variants of Transformers: Transformer~\cite{NIPS2017_3f5ee243}, Informer~\cite{haoyietal-informer-2021}, Reformer~\cite{kitaev2020reformer} and Autoformer~\cite{wu2021autoformer} to validate the generality of our framework.

\paragraph{Implementation details} All the experiments are implemented with PyTorch~\cite{Paszke2019PyTorchAI} and conducted on a single NVIDIA TITAN V 12GB GPU. Each model is trained by ADAM~\cite{DBLP:journals/corr/KingmaB14} using L2 loss with the initial learning rate of $10^{-4}$ and batch size of 32. Each Transformer-based model contains two encoder layers and one decoder layer. Considering the efficiency of hyperparameters search, we use two-layer perceptron projector with the hidden dimension varying in $\{64, 128, 256\}$ in De-stationary Attention. We repeat each experiment three times with different random seeds and report the test MSE/MAE under different prediction lengths, and the standard deviations are also provided in the Appendix~\ref{app:full}. A lower MSE/MAE indicates better performance. 

\subsection{Main Results}
\paragraph{Forecasting results} As for multivariate forecasting results, the vanilla Transformer equipped with our framework consistently achieves state-of-the-art performance in all benchmarks and prediction lengths (Table~\ref{tab:results}). Notably, Non-stationary Transformer outperforms other deep models impressively on datasets characterized by high non-stationarity: under the prediction length of 336, we achieve \textbf{17\%} MSE reduction~($0.509\to0.421$) on Exchange and \textbf{25\%}~($2.669\to2.010$) on ILI compared to previous state-of-the-art results, which indicates that the potential of deep model is still constrained on non-stationary data. We also list the univariate results of two typical datasets with different stationarity in Table~\ref{tab:uni_results}. Non-stationary Transformer still realizes remarkable forecasting performance.

\begin{table}[htbp]
  \caption{Forecasting results comparison under different prediction lengths $O \in \{96,192,336,720\}$. The input sequence length is set to 36 for ILI and 96 for the others. Additional results (ETTm1, ETTh1, ETTh2) can be found in Appendix~\ref{app:add}.}
  \label{tab:results}
  \centering
  \begin{threeparttable}
  \begin{small}
  \renewcommand{\multirowsetup}{\centering}
  \setlength{\tabcolsep}{2pt}
  \begin{tabular}{c|c|cccccccccccccc}
    \toprule
    \multicolumn{2}{c}{Models} & \multicolumn{2}{c}{\textbf{Ours}} & \multicolumn{2}{c}{\scalebox{0.8}{Autoformer~\cite{wu2021autoformer}}} & \multicolumn{2}{c}{\scalebox{0.8}{Pyraformer~\cite{liu2021pyraformer}}} &  \multicolumn{2}{c}{Informer~\cite{haoyietal-informer-2021}} & \multicolumn{2}{c}{LogTrans~\cite{2019Enhancing}}  & \multicolumn{2}{c}{Reformer~\cite{kitaev2020reformer}} & \multicolumn{2}{c}{LSTNet~\cite{2018Modeling}}  \\
    \cmidrule(lr){3-4} \cmidrule(lr){5-6}\cmidrule(lr){7-8} \cmidrule(lr){9-10}\cmidrule(lr){11-12}\cmidrule(lr){13-14}\cmidrule(lr){15-16}
    \multicolumn{2}{c}{Metric} & MSE & MAE & MSE & MAE & MSE & MAE & MSE & MAE & MSE & MAE & MSE & MAE & MSE & MAE  \\
    \toprule
    \multirow{4}{*}{\rotatebox{90}{Exchange}}
    &  96 & \textbf{0.111} & \textbf{0.237} & 0.197 & 0.323 & 0.852 & 0.780 & 0.847 & 0.752 & 0.968 & 0.812 & 1.065 & 0.829 & 1.551 & 1.058   \\
    & 192 & \textbf{0.219} & \textbf{0.335} & 0.300 & 0.369 & 0.993 & 0.858 & 1.204 & 0.895 & 1.040 & 0.851 & 1.188 & 0.906 & 1.477 & 1.028   \\
    & 336 & \textbf{0.421} & \textbf{0.476} & 0.509 & 0.524 & 1.240 & 0.958 & 1.672 & 1.036 & 1.659 & 1.081 & 1.357 & 0.976 & 1.507 & 1.031  \\
    & 720 & \textbf{1.092} & \textbf{0.769} & 1.447 & 0.941 & 1.711 & 1.093 & 2.478 & 1.310 & 1.941 & 1.127 & 1.510 & 1.016 & 2.285 & 1.243  \\
    \midrule
    \multirow{4}{*}{\rotatebox{90}{ILI}}  
    & 24   & \textbf{2.294} & \textbf{0.945} & 3.483 & 1.287 & 5.800 & 1.693 & 5.764 & 1.677 & 4.480 & 1.444 & 4.400 & 1.382 & 6.026 & 1.770   \\
    & 36   & \textbf{1.825} & \textbf{0.848} & 3.103 & 1.148 & 6.043 & 1.733 & 4.755 & 1.467 & 4.799 & 1.467 & 4.783 & 1.448 & 5.340 & 1.668  \\
    & 48   & \textbf{2.010} & \textbf{0.900} & 2.669 & 1.085 & 6.213 & 1.763 & 4.763 & 1.469 & 4.800 & 1.468 & 4.832 & 1.465 & 6.080 & 1.787   \\
    & 60   & \textbf{2.178} & \textbf{0.963} & 2.770 & 1.125 & 6.531 & 1.814 & 5.264 & 1.564 & 5.278 & 1.560 & 4.882 & 1.483 & 5.548 & 1.720   \\
    \midrule
    \multirow{4}{*}{\rotatebox{90}{ETTm2}} 
    &  96 & \textbf{0.192} & \textbf{0.274} & 0.255 & 0.339 & 0.409 & 0.488 & 0.365 & 0.453 & 0.768 & 0.642 & 0.658 & 0.619 & 3.142 & 1.365  \\
    & 192 & \textbf{0.280} & \textbf{0.339} & 0.281 & 0.340 & 0.673 & 0.641 & 0.533 & 0.563 & 0.989 & 0.757 & 1.078 & 0.827 & 3.154 & 1.369  \\
    & 336 & \textbf{0.334} & \textbf{0.361} & 0.339 & 0.372 & 1.210 & 0.846 & 1.363 & 0.887 & 1.334 & 0.872 & 1.549 & 0.972 & 3.160 & 1.369 \\
    & 720 & \textbf{0.417} & \textbf{0.413} & 0.422 & 0.419 & 4.044 & 1.526 & 3.379 & 1.388 & 3.048 & 1.328 & 2.631 & 1.242 & 3.171 & 1.368   \\
    \midrule
    \multirow{4}{*}{\rotatebox{90}{Electricity}}
    &  96  & \textbf{0.169} & \textbf{0.273} & 0.201 & 0.317 & 0.498 & 0.299 & 0.274 & 0.368 & 0.258 & 0.357 & 0.312 & 0.402 & 0.680 & 0.645   \\
    & 192  & \textbf{0.182} & \textbf{0.286} & 0.222 & 0.334 & 0.828 & 0.312 & 0.296 & 0.386 & 0.266 & 0.368 & 0.348 & 0.433 & 0.725 & 0.676  \\
    & 336  & \textbf{0.200} & \textbf{0.304} & 0.231 & 0.338 & 1.476 & 0.326 & 0.300 & 0.394 & 0.280 & 0.380 & 0.350 & 0.433 & 0.828 & 0.727   \\
    & 720  & \textbf{0.222} & \textbf{0.321} & 0.254 & 0.361 & 4.090 & 0.372 & 0.373 & 0.439 & 0.283 & 0.376 & 0.340 & 0.420 & 0.957 & 0.811  \\
    \midrule
    \multirow{4}{*}{\rotatebox{90}{Traffic}} 
    &  96 & \textbf{0.612} & \textbf{0.338} & 0.613 & 0.388 & 0.684 & 0.393 & 0.719 & 0.391 & 0.684 & 0.384 & 0.732 & 0.423 & 1.107 & 0.685  \\
    & 192 & \textbf{0.613} & \textbf{0.340} & 0.616 & 0.382 & 0.692 & 0.394 & 0.696 & 0.379 & 0.685 & 0.390 & 0.733 & 0.420 & 1.157 & 0.706   \\
    & 336 & \textbf{0.618} & \textbf{0.328} & 0.622 & 0.337 & 0.699 & 0.396 & 0.777 & 0.420 & 0.733 & 0.408 & 0.742 & 0.420 & 1.216 & 0.730  \\
    & 720 & \textbf{0.653} & \textbf{0.355} & 0.660 & 0.408 & 0.712 & 0.404 & 0.864 & 0.472 & 0.717 & 0.396 & 0.755 & 0.423 & 1.481 & 0.805  \\
    \midrule
    \multirow{4}{*}{\rotatebox{90}{Weather}}  
    &  96 & \textbf{0.173} & \textbf{0.223} & 0.266 & 0.336 & 0.354 & 0.392 & 0.300 & 0.384 & 0.458 & 0.490 & 0.689 & 0.596 & 0.594 & 0.587  \\
    & 192 & \textbf{0.245} & \textbf{0.285} & 0.307 & 0.367 & 0.673 & 0.597 & 0.598 & 0.544 & 0.658 & 0.589 & 0.752 & 0.638 & 0.560 & 0.565  \\
    & 336 & \textbf{0.321} & \textbf{0.338} & 0.359 & 0.395 & 0.634 & 0.592 & 0.578 & 0.523 & 0.797 & 0.652 & 0.639 & 0.596 & 0.597 & 0.587  \\
    & 720 & \textbf{0.414} & \textbf{0.410} & 0.419 & 0.428 & 0.942 & 0.723 & 1.059 & 0.741 & 0.869 & 0.675 & 1.130 & 0.792 & 0.618 & 0.599  \\
    \bottomrule
  \end{tabular}
  \end{small}
  \end{threeparttable}
  \vspace{-10pt}
\end{table}

\begin{table}[htbp]
  \caption{Univariate results under different prediction lengths $O \in \{96,192,336,720\}$ on two typical datasets with strong non-stationary. The input sequence length is set to 96.}
  \label{tab:uni_results}
  \centering
  \begin{small}
  \renewcommand{\multirowsetup}{\centering}
  \setlength{\tabcolsep}{1.6pt}
  \begin{tabular}{c|c|ccccccccccccccccc}
    \toprule
    \multicolumn{2}{c}{Models} & \multicolumn{2}{c}{\scalebox{0.9}{\textbf{Ours} }} &
    \multicolumn{2}{c}{\scalebox{0.90}{N-HiTS~\cite{challu2022n}}}  & 
    \multicolumn{2}{c}{\scalebox{0.80}{N-BEATS~\cite{oreshkin2019n}}} &
    \multicolumn{2}{c}{\scalebox{0.75}{Autoformer~\cite{wu2021autoformer}}}  & 
    \multicolumn{2}{c}{\scalebox{0.75}{Pyraformer~\cite{liu2021pyraformer}}} &
    \multicolumn{2}{c}{\scalebox{0.90}{Informer~\cite{haoyietal-informer-2021}}} & 
    \multicolumn{2}{c}{\scalebox{0.90}{Reformer~\cite{kitaev2020reformer}}} & 
 \multicolumn{2}{c}{\scalebox{0.90}{ARIMA~\cite{Anderson1976TimeSeries2E}}}  \\
    \cmidrule(lr){3-4} \cmidrule(lr){5-6}\cmidrule(lr){7-8} \cmidrule(lr){9-10}\cmidrule(lr){11-12}\cmidrule(lr){13-14}\cmidrule(lr){15-16}\cmidrule(lr){17-18}
    \multicolumn{2}{c}{Metric} & \scalebox{0.90}{MSE} & \scalebox{0.90}{MAE} & \scalebox{0.90}{MSE} & \scalebox{0.90}{MAE} & \scalebox{0.90}{MSE} & \scalebox{0.90}{MAE} & \scalebox{0.90}{MSE} & \scalebox{0.90}{MAE} & \scalebox{0.90}{MSE} & \scalebox{0.90}{MAE} & \scalebox{0.90}{MSE} & \scalebox{0.90}{MAE} & \scalebox{0.90}{MSE} & \scalebox{0.90}{MAE} & \scalebox{0.90}{MSE} & \scalebox{0.90}{MAE}  \\
    \toprule
    \multirow{4}{*}{\scalebox{0.90}{\rotatebox{90}{Exchange}}} 
    & \scalebox{0.90}{96}
    & \scalebox{0.90}{\textbf{0.104}} & \scalebox{0.90}{\textbf{0.235}} 
    & \scalebox{0.90}{0.114} & \scalebox{0.90}{0.248} 
    & \scalebox{0.90}{0.156} & \scalebox{0.90}{0.299} 
    & \scalebox{0.90}{0.241} & \scalebox{0.90}{0.387} 
    & \scalebox{0.90}{0.290} & \scalebox{0.90}{0.439} 
    & \scalebox{0.90}{0.591} & \scalebox{0.90}{0.615} 
    & \scalebox{0.90}{1.327} & \scalebox{0.90}{0.944} 
    & \scalebox{0.90}{0.112} & \scalebox{0.90}{0.245} \\
    
    & \scalebox{0.90}{192} 
    & \scalebox{0.90}{\textbf{0.230}} & \scalebox{0.90}{\textbf{0.375}} 
    & \scalebox{0.90}{0.250} & \scalebox{0.90}{0.387} 
    & \scalebox{0.90}{0.669} & \scalebox{0.90}{0.665} 
    & \scalebox{0.90}{0.273} & \scalebox{0.90}{0.403}
    & \scalebox{0.90}{0.594} & \scalebox{0.90}{0.644} 
    & \scalebox{0.90}{1.183} & \scalebox{0.90}{0.912} 
    & \scalebox{0.90}{1.258} & \scalebox{0.90}{0.924} 
    & \scalebox{0.90}{0.304} & \scalebox{0.90}{0.404} \\
    
    & \scalebox{0.90}{336} 
    & \scalebox{0.90}{\textbf{0.432}} & \scalebox{0.90}{\textbf{0.509}} 
    & \scalebox{0.90}{0.434} & \scalebox{0.90}{0.516} 
    & \scalebox{0.90}{0.611} & \scalebox{0.90}{0.605} 
    & \scalebox{0.90}{0.508} & \scalebox{0.90}{0.539} 
    & \scalebox{0.90}{0.962} & \scalebox{0.90}{0.824} 
    & \scalebox{0.90}{1.367} & \scalebox{0.90}{0.984} 
    & \scalebox{0.90}{2.179} & \scalebox{0.90}{1.296} 
    & \scalebox{0.90}{0.736} & \scalebox{0.90}{0.598} \\
    
    & \scalebox{0.90}{720} 
    & \scalebox{0.90}{\textbf{0.782}} & \scalebox{0.90}{\textbf{0.682}} 
    & \scalebox{0.90}{1.061} & \scalebox{0.90}{0.773} 
    & \scalebox{0.90}{1.111} & \scalebox{0.90}{0.860}    
    & \scalebox{0.90}{0.991} & \scalebox{0.90}{0.768} 
    & \scalebox{0.90}{1.285} & \scalebox{0.90}{0.958}
    & \scalebox{0.90}{1.872} & \scalebox{0.90}{1.072} 
    & \scalebox{0.90}{1.280} & \scalebox{0.90}{0.953} 
    & \scalebox{0.90}{1.871} & \scalebox{0.90}{0.935} \\
    
    \midrule
    \multirow{4}{*}{\scalebox{0.90}{\rotatebox{90}{ETTm2}}}
    & \scalebox{0.90}{96} 
    & \scalebox{0.90}{0.069} & \scalebox{0.90}{0.193} 
    & \scalebox{0.90}{0.092} & \scalebox{0.90}{0.232} 
    & \scalebox{0.90}{0.082} & \scalebox{0.90}{0.219}     
    & \scalebox{0.90}{\textbf{0.065}} & \scalebox{0.90}{\textbf{0.189}}
    & \scalebox{0.90}{0.074} & \scalebox{0.90}{0.208}
    & \scalebox{0.90}{0.088} & \scalebox{0.90}{0.225}
    & \scalebox{0.90}{0.131} & \scalebox{0.90}{0.288}  
    & \scalebox{0.90}{0.211} & \scalebox{0.90}{0.362}  \\
    
    & \scalebox{0.90}{192} 
    & \scalebox{0.90}{\textbf{0.109}} & \scalebox{0.90}{\textbf{0.249}} 
    & \scalebox{0.90}{0.128} & \scalebox{0.90}{0.276}
    & \scalebox{0.90}{0.120} & \scalebox{0.90}{0.268}    
    & \scalebox{0.90}{0.118} & \scalebox{0.90}{0.256}  
    & \scalebox{0.90}{0.116} & \scalebox{0.90}{0.252}  
    & \scalebox{0.90}{0.132} & \scalebox{0.90}{0.283} 
    & \scalebox{0.90}{0.186} & \scalebox{0.90}{0.354} 
    & \scalebox{0.90}{0.261} & \scalebox{0.90}{0.406} \\
    
    & \scalebox{0.90}{336}  
    & \scalebox{0.90}{\textbf{0.139}} & \scalebox{0.90}{\textbf{0.286}} 
    & \scalebox{0.90}{0.165} & \scalebox{0.90}{0.314} 
    & \scalebox{0.90}{0.226} & \scalebox{0.90}{0.370} 
    & \scalebox{0.90}{0.154} & \scalebox{0.90}{0.305} 
    & \scalebox{0.90}{0.143} & \scalebox{0.90}{0.295} 
    & \scalebox{0.90}{0.180} & \scalebox{0.90}{0.336} 
    & \scalebox{0.90}{0.220} & \scalebox{0.90}{0.381} 
    & \scalebox{0.90}{0.317} & \scalebox{0.90}{0.448} \\
    
    & \scalebox{0.90}{720} 
    & \scalebox{0.90}{\textbf{0.180}} & \scalebox{0.90}{\textbf{0.331}} 
    & \scalebox{0.90}{0.243} & \scalebox{0.90}{0.397} 
    & \scalebox{0.90}{0.188} & \scalebox{0.90}{0.338} 
    & \scalebox{0.90}{0.182} & \scalebox{0.90}{0.335} 
    & \scalebox{0.90}{0.197} & \scalebox{0.90}{0.338} 
    & \scalebox{0.90}{0.300} & \scalebox{0.90}{0.435} 
    & \scalebox{0.90}{0.267} & \scalebox{0.90}{0.430} 
    & \scalebox{0.90}{0.366} & \scalebox{0.90}{0.487}  \\
    \bottomrule
  \end{tabular}
  \end{small}
  \vspace{-5pt}
\end{table}

\paragraph{Framework generality}  We apply our framework to four mainstream Transformers and report the performance promotion of each model (Table~\ref{tab:boost}). Our method consistently improves the forecasting ability of different models. Overall, it achieves averaged \textbf{49.43\%} promotion on Transformer, \textbf{47.34\%} on Informer, \textbf{46.89\%} on Reformer and \textbf{10.57\%} on Autoformer, making each of them surpass previous state-of-the-art. Compared to native blocks of the models, there is hardly any parameter and computation increase by applying our framework (See 
Appendix~\ref{app:eff} for details), and thereby their computational complexities can be preserved. It validates that Non-stationary Transformer is an effective and lightweight framework that can be widely applied to Transformer-based models and enhances their non-stationary predictability to achieve state-of-the-art performance.

\begin{table}[tbp]
  \caption{Performance promotion by applying the proposed framework to Transformer and its variants. We report the averaged MSE/MAE of all prediction lengths (stated in Table~\ref{tab:results}) and the relative MSE reduction ratios (Promotion) by our framework. Full results (under all prediction lengths and promotion on  ETSformer~\cite{woo2022etsformer}, FEDformer~\cite{zhou2022fedformer}) can be found in Appendix~\ref{app:full}.}
  \label{tab:boost}
  \centering
  \begin{threeparttable}
  \begin{small}
  \renewcommand{\multirowsetup}{\centering}
  \setlength{\tabcolsep}{4.2pt}
  \begin{tabular}{c|cc|cc|cc|cc|cc|cccc}
    \toprule
    Dataset & \multicolumn{2}{c}{Exchange} & \multicolumn{2}{c}{ILI} & \multicolumn{2}{c}{ETTm2}  & \multicolumn{2}{c}{Electricity} & \multicolumn{2}{c}{Traffic} & \multicolumn{2}{c}{Weather}  \\
    \cmidrule(lr){2-3} \cmidrule(lr){4-5}\cmidrule(lr){6-7} \cmidrule(lr){8-9}\cmidrule(lr){10-11}\cmidrule(lr){12-13}
    Model & MSE & MAE & MSE & MAE & MSE & MAE & MSE & MAE & MSE & MAE & MSE & MAE  \\
    \toprule
    Transformer         & 1.425 & 0.915 & 4.864 & 1.460 & 1.501 & 0.869 & 0.277 & 0.372 & 0.665 & 0.363 & 0.657 & 0.573 \\
    \textbf{ + Ours }            & \textbf{0.457} & \textbf{0.449} & \textbf{2.077} & \textbf{0.914} & \textbf{0.306} & \textbf{0.347} & \textbf{0.193} & \textbf{0.296} & \textbf{0.628} & \textbf{0.345} & \textbf{0.288} & \textbf{0.314} \\
    \cmidrule(lr){1-13}
    Promotion & \multicolumn{2}{c|}{67.93\%} & \multicolumn{2}{c|}{57.30\%} & \multicolumn{2}{c|}{79.61\%} & \multicolumn{2}{c|}{30.32\%} & \multicolumn{2}{c|}{5.56\%} & \multicolumn{2}{c}{56.16\%} \\
    \midrule
    Informer            & 1.550 & 0.998 & 5.137 & 1.544 & 1.410 & 0.823  & 0.311 & 0.397 & 0.764 & 0.416 & 0.634 & 0.548 \\
    \textbf{ + Ours }             & \textbf{0.496} & \textbf{0.460} & \textbf{2.125} & \textbf{0.928} & \textbf{0.460} & \textbf{0.434} & \textbf{0.226} & \textbf{0.330}  & \textbf{0.719} & \textbf{0.409} & \textbf{0.275} & \textbf{0.302} \\
    \cmidrule(lr){1-13}
    Promotion & \multicolumn{2}{c|}{68.00\%} & \multicolumn{2}{c|}{58.63\%} & \multicolumn{2}{c|}{67.38\%} & \multicolumn{2}{c|}{27.33\%} & \multicolumn{2}{c|}{5.89\%} & \multicolumn{2}{c}{56.78\%} \\
    \midrule
    Reformer            & 1.280 & 0.932 & 4.724 & 1.443 & 1.479 & 0.915 & 0.338 & 0.429 & 0.741 & 0.423 & 0.803 & 0.656\\
    \textbf{ + Ours }             & \textbf{0.462} & \textbf{0.468} & \textbf{2.865} & \textbf{1.065} & \textbf{0.493} & \textbf{0.441} & \textbf{0.206} & \textbf{0.308} & \textbf{0.682} & \textbf{0.372} & \textbf{0.286} & \textbf{0.308} \\
    \cmidrule(lr){1-13}
    Promotion & \multicolumn{2}{c|}{63.91\%} & \multicolumn{2}{c|}{39.35\%} & \multicolumn{2}{c|}{66.67\%} & \multicolumn{2}{c|}{39.05\%} & \multicolumn{2}{c|}{7.96\%} & \multicolumn{2}{c}{64.38\%} \\
    \midrule
    Autoformer     & 0.613          & 0.539          & 3.006          & 1.161                  & 0.324          & 0.368   & 0.227          & 0.338         & 0.628 
    & 0.379        & 0.338          & 0.382  \\
    \textbf{ + Ours }        & \textbf{0.487} & \textbf{0.491} & \textbf{2.545} & \textbf{1.039} & \textbf{0.305} & \textbf{0.345}  & \textbf{0.216} & \textbf{0.315} & \textbf{0.619} &  \textbf{0.364} & \textbf{0.286} & \textbf{0.310} \\
    \cmidrule(lr){1-13}
    Promotion & \multicolumn{2}{c|}{20.55\%} & \multicolumn{2}{c|}{15.34\%} & \multicolumn{2}{c|}{5.86\%}  & \multicolumn{2}{c|}{4.85\%}& \multicolumn{2}{c|}{1.43\%} & \multicolumn{2}{c}{15.38\%} \\
    \bottomrule
  \end{tabular}
  \end{small}
  \end{threeparttable}
  \vspace{-10pt}
\end{table}

\subsection{Ablation Study}
\begin{figure}[htbp]
  \includegraphics[width=\columnwidth]{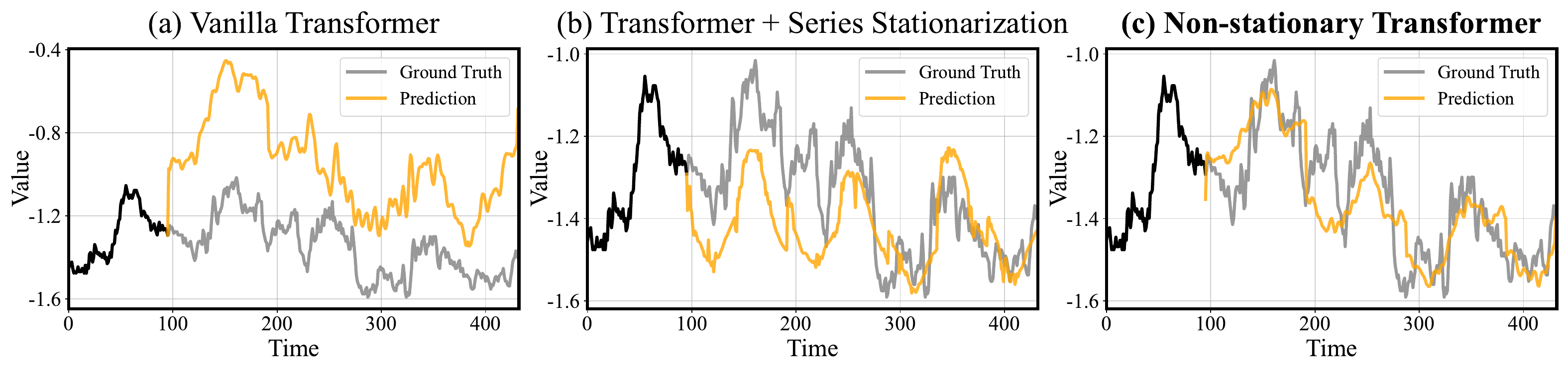}
  \centering
  \vspace{-23pt}
  \caption{Visualization of ETTm2 predictions given by different models.}
  \label{fig:case}
  \vspace{-10pt}
\end{figure}

\paragraph{Quality evaluation} To explore the role of each module in our proposed framework, we compare the prediction results on ETTm2 obtained by three models: vanilla Transformer, Transformer with only Series Stationarization, and our Non-stationary Transformer. In Figure~\ref{fig:case}, we find out that the two modules strengthen the non-stationary forecasting ability of Transformer from different perspectives. Series Stationarization focuses on the alignment of statistical properties among each series input that benefits Transformer a lot to generalize on out-of-distribution data. However, as is shown in Figure~\ref{fig:case}(b), the over-stationarized circumstance for training makes the deep model more likely to output uneventful series with significant high stationarity and neglect the nature of non-stationary real-world data. With the aid of De-stationary Attention, the model gives concern back to the inherent non-stationarity of real-world time series. It is beneficial for an accurate prediction of the detailed series variation, which is vital in real-world time series forecasting.

\vspace{-5pt}

\paragraph{Quantitative performance} In addition to the above case study, we also provide quantitative forecasting performance comparison with stationarization methods: a deep method RevIN~\cite{kim2022reversible} and Series Stationarization (Section~\ref{sec:sm}). As is shown in Table~\ref{tab:comparison}, the forecasting results assisted by RevIN and Series Stationarization are basically the same, which indicates that the parameter-free version of normalization in our framework performs sufficiently to stationarize time series. Besides, the proposed De-stationary Attention in Non-stationary Transformers further boosts the performance and achieves the best in all six benchmarks. The MSE reduction brought by De-stationary Attention becomes significant, especially when the dataset is highly non-stationary (Exchange: $0.569\to 0.461$, ETTm2: $0.461\to 0.306$). The comparison reveals that simply stationarizing time series still limits the predictive capability of Transformers, and the complementary mechanisms in Non-stationary Transformers can properly release the models' potential for non-stationary series forecasting.
\begin{table}[tbp]
  \caption{Forecasting results obtained by applying different methods to Transformer and Reformer. We report the averaged MSE/MAE of all prediction lengths (stated in Table~\ref{tab:results}) for comparison. Complete results can be found in Appendix~\ref{app:comp}.}
  \label{tab:comparison}
  \centering
  \begin{threeparttable}
  \begin{small}
  \renewcommand{\multirowsetup}{\centering}
  \setlength{\tabcolsep}{4.0pt}
  \begin{tabular}{c|c|cccccc|cccccc}
    \toprule
    \multicolumn{2}{c}{Base Models} &
    \multicolumn{6}{c}{ Transformer } &
    \multicolumn{6}{c}{ Reformer }  \\
    \cmidrule(lr){3-8}  \cmidrule(lr){9-14}
    
    \multicolumn{2}{c}{\multirow{2}{*}{Methods}} &
    \multicolumn{2}{c}{\multirow{2}{*}{+ RevIN~\cite{kim2022reversible}}} &
    \multicolumn{2}{c}{\scalebox{0.85}{+ Series}} &
    \multicolumn{2}{c}{\multirow{2}{*}{+ \textbf{Ours}}} &
    \multicolumn{2}{c}{\multirow{2}{*}{+ RevIN~\cite{kim2022reversible}}} & 
    \multicolumn{2}{c}{\scalebox{0.85}{+ Series}} & 
    \multicolumn{2}{c}{\multirow{2}{*}{+ \textbf{Ours}}} \\
    
    \multicolumn{2}{c}{ } & \multicolumn{2}{c}{ }  & \multicolumn{2}{c}{\scalebox{0.85}{Stationarization}} & \multicolumn{2}{c}{ } & \multicolumn{2}{c}{ } & \multicolumn{2}{c}{\scalebox{0.85}{Stationarization}} & \multicolumn{2}{c}{ }  \\
    \cmidrule(lr){3-4} \cmidrule(lr){5-6}
    \cmidrule(lr){7-8} \cmidrule(lr){9-10}
    \cmidrule(lr){11-12} \cmidrule(lr){13-14}
    \multicolumn{2}{c}{Metric} & MSE & MAE & MSE & MAE & MSE & MAE & MSE & MAE & MSE & MAE & MSE & MAE \\
    \toprule
    \multicolumn{2}{c}{Exchange}    &	0.567&	0.487&0.569&	0.488&	\textbf{0.461}&	\textbf{0.454}&	0.469&	0.472&	0.470&	0.473&	\textbf{0.462}&	\textbf{0.468}\\
    \midrule
    \multicolumn{2}{c}{ILI}         &	2.205&	0.934&2.206&	0.934&	\textbf{2.077}&	\textbf{0.914}&	3.024&	1.096&	3.023&	1.096&	\textbf{2.865}&	\textbf{1.065}\\
    \midrule
    \multicolumn{2}{c}{ETTm2}       &	0.460&	0.416&0.461&	0.416&	\textbf{0.306}&	\textbf{0.347}&	0.542&	0.459&	0.537&	0.459&	\textbf{0.493}&	\textbf{0.441}\\
    \midrule
    \multicolumn{2}{c}{Electricity} &	0.197&	0.298&0.197&	0.298&	\textbf{0.193}&	\textbf{0.296}&	0.208&	0.309&	0.207&	0.309&	\textbf{0.206}&	\textbf{0.308}\\
    \midrule
    \multicolumn{2}{c}{Traffic}     &	0.643&	0.352&0.641&	0.352&	\textbf{0.628}&	\textbf{0.345}&	0.687&	0.378&	0.691&	0.380&	\textbf{0.682}&	\textbf{0.372}\\
    \midrule
    \multicolumn{2}{c}{Weather}     &	0.301&	0.316&0.304&	0.317&	\textbf{0.288}&	\textbf{0.314}&	0.291&	0.309&	0.292&	0.309&	\textbf{0.286}&	\textbf{0.308}\\
    \bottomrule
  \end{tabular}
  \end{small}
  \end{threeparttable}
  \vspace{-5pt}
\end{table}

\subsection{Model Analysis}

\begin{figure}[htbp]
  \includegraphics[width=\columnwidth]{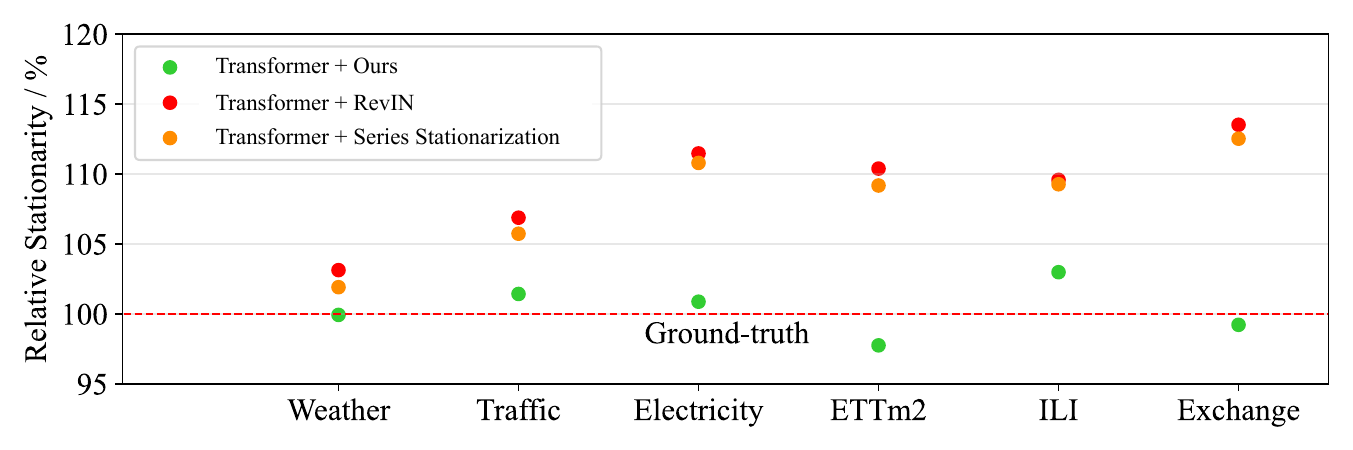}
  \centering
  \vspace{-15pt}
  \caption{Relative stationarity is calculated as the ratio of ADF test statistics between the model predictions and ground truth. From left to right, the dataset is increasingly non-stationary. While models equipped with only stationarization tend to output highly stationary series, our method gives predictions with stationarity closer to ground truth.}
  \label{fig:stationarities}
  \vspace{-5pt}
\end{figure}

\paragraph{Over-stationarization problem} To verify the over-stationarization problem from a statistical view, we train Transformers with the aforementioned methods respectively, arrange all predicted time series in chronological order and compare the degree of stationarity with the ground truth (Figure~\ref{fig:stationarities}). While models solely equipped with stationarization methods tend to output series with unexpected high degree of stationarity, the results assisted by De-stationary Attention are close to the actual value (relative stationarity $\in [97\%, 103\%]$). Besides, as the degree of series stationarity increases, the over-stationarization problem becomes more significant. The huge discrepancy of the degree of stationarity can account for the inferior performance of Transformer with only stationarization. And it also demonstrates that De-stationary Attention as an internal renovation alleviates over-stationarization.

\paragraph{Exploring of Non-stationary Information Re-incorporation}
It is notable that by specifying over-stationarization as less distinguishable attention, we narrow down our design space into the attention calculation mechanism. To explore other approaches to retrieve non-stationary information, we conduct experiments by re-incorporating the $\mu$ and $\sigma$ into feed-forward layers (DeFF), which is the left part of the Transformer architecture. In detail, we feed learned $\mu$ and $\sigma$ into each feed-forward layer iteratively. As is shown in Table~\ref{tab:re-incorporate}, re-incorporating non-stationarity is necessary only when the inputs are stationarized (Stationary), which is beneficial for forecasting but will lead to stationarity discrepancy of model outputs. And our proposed design (Stat + DeAttn) makes further promotion and achieves the best in most cases (77\%). In addition to the theoretical analysis, experimental results further validate the effectiveness of our design in re-incorporating non-stationarity on attention.

\begin{table}[htbp]
  \caption{Ablation of framework design. \emph{Baseline} means vanilla Transformer, \emph{Stationary} means adding Series Stationarization, \emph{DeFF} means re-incorporating non-stationarity on feed-forward layers, \emph{DeAttn} means re-incorporating by De-stationary Attention, \emph{Stat + DeFF} means adding Series Stationarization and re-incorporating on feed-forward layers. \emph{Stat + DeAttn} means our proposed framework. 
  }
  \label{tab:re-incorporate}
  \centering
  \begin{small}
  \renewcommand{\multirowsetup}{\centering}
  \setlength{\tabcolsep}{4.6pt}
  \begin{tabular}{c|c|cc|cc|cc|cc|cc|cc}
    \toprule
    \multicolumn{2}{c}{Models} &
    \multicolumn{2}{c}{Baseline} &
    \multicolumn{2}{c}{Stationary} &
    \multicolumn{2}{c}{DeFF} &
    \multicolumn{2}{c}{DeAttn} &
    \multicolumn{2}{c}{Stat + DeFF} & 
    \multicolumn{2}{c}{Stat + DeAttn} \\
    \cmidrule(lr){3-4} 
    \cmidrule(lr){5-6}
    \cmidrule(lr){7-8} 
    \cmidrule(lr){9-10}
    \cmidrule(lr){11-12}
    \cmidrule(lr){13-14}
    \multicolumn{2}{c}{Metric} & MSE & MAE & MSE & MAE & MSE & MAE & MSE & MAE & MSE & MAE & MSE & MAE \\
    \toprule
    \multirow{4}{*}{\rotatebox{90}{Exchange}} 
    & 96  & 0.567 & 0.591 & 0.136 & 0.258 & 0.784 & 0.696 & 0.611 & 0.613 & 0.116 & 0.243 & \textbf{0.111} & \textbf{0.237}\\
    & 192 & 1.150 & 0.825 & 0.239 & 0.348 & 1.162 & 0.866 & 1.202 & 0.840 & 0.280 & 0.383 & \textbf{0.219} & \textbf{0.335}\\
    & 336 & 1.792 & 1.084 & 0.425 & 0.479 & 1.346 & 0.963 & 1.516 & 0.981 & \textbf{0.371} & \textbf{0.452} & 0.421 & 0.476\\
    & 720 & 2.191 & 1.159 & 1.475 & 0.865 & 2.042 & 1.163 & 2.894 & 1.377 & \textbf{0.934} & \textbf{0.704} & 1.092 & 0.769\\
    \midrule
    \multirow{4}{*}{\rotatebox{90}{ILI}} 
    & 24 & 4.748 & 1.430 & 2.573 & 0.980 & 4.850 & 1.445 & 4.734 & 1.424 & 2.404 & 0.985 & \textbf{2.294} & \textbf{0.945}\\
    & 36 & 4.671 & 1.430 & 1.955 & 0.870 & 4.848 & 1.452 & 4.927 & 1.482 & 2.585 & 0.983 & \textbf{1.825} & \textbf{0.848}\\
    & 48 & 4.994 & 1.482 & 2.057 & 0.902 & 4.903 & 1.466 & 4.996 & 1.483 & 2.496 & 0.991 & \textbf{2.010} & \textbf{0.900}\\
    & 60 & 5.041 & 1.499 & 2.238 & 0.982 & 5.196 & 1.524 & 5.184 & 1.519 & 2.667 & 1.059 & \textbf{2.178} & \textbf{0.963}\\
    \midrule
    \multirow{4}{*}{\rotatebox{90}{ETTm2}} 
    & 96 &  0.572 & 0.552 & 0.253 & 0.311 & 0.767 & 0.635 & 0.304 & 0.406 & 0.275 & 0.329 & \textbf{0.192} & \textbf{0.274}\\
    & 192 & 1.161 & 0.793 & 0.453 & 0.404 & 0.960 & 0.717 & 0.820 & 0.652 & 0.406 & 0.403 & \textbf{0.280} & \textbf{0.339}\\
    & 336 & 1.209 & 0.842 & 0.546 & 0.461 & 1.159 & 0.811 & 1.406 & 0.883 & 0.502 & 0.465 & \textbf{0.334} & \textbf{0.361} \\
    & 720 & 3.061 & 1.289 & 0.593 & 0.489 & 3.187 & 1.308 & 2.858 & 1.108 & 0.694 & 0.575 & \textbf{0.417} & \textbf{0.413}\\
    \midrule
    \multirow{4}{*}{\rotatebox{90}{Electricity}} 
    & 96  &  0.260 & 0.358 &  0.171 & 0.275 & 0.260 & 0.356 & 0.253 & 0.351 & 0.170 & 0.274 & \textbf{0.169}   &  \textbf{0.273}\\
    & 192 &  0.266 & 0.367 &  0.192 & 0.296 & 0.264 & 0.365 & 0.257 & 0.358 & 0.188 & 0.293 & \textbf{0.182} & \textbf{0.286} \\
    & 336 &  0.280 & 0.375 &  0.208 & 0.306 & 0.277 & 0.374 & 0.270 & 0.365 & 0.206 & 0.309 & \textbf{0.200} & \textbf{0.304}\\
    & 720 &  0.302 & 0.386 &  \textbf{0.216} &  \textbf{0.315} & 0.299 & 0.384 & 0.295 & 0.380  & 0.223 &  0.323 &  0.222 & 0.321\\
    \midrule
    \multirow{4}{*}{\rotatebox{90}{Traffic}} 
    & 96  & 0.647 & 0.357 & 0.614 & 0.337 & 0.646 & 0.353 & 0.650 & 0.358 & \textbf{0.605} & \textbf{0.333} & 0.612 & 0.338 \\
    & 192 & 0.649 & 0.356 & 0.637 & 0.351 & 0.645 & 0.352 & 0.655 & 0.358 & 0.617 & 0.342 & \textbf{0.613} & \textbf{0.340} \\
    & 336 & 0.667 & 0.364 & 0.653 & 0.359 & 0.672 & 0.360 & 0.656 & 0.355 & 0.635 & 0.349 & \textbf{0.618} & \textbf{0.328} \\
    & 720 & 0.697 & 0.376 & 0.661 & 0.360 & 0.695 & 0.376 & 0.681 & 0.366 & \textbf{0.649} & \textbf{0.351} & 0.653 & 0.355 \\
    \midrule
    \multirow{4}{*}{\rotatebox{90}{Weather}} 
    & 96  & 0.395 & 0.427 & 0.175 & 0.225 & 0.417 & 0.445 & 0.296 & 0.364 & 0.178 & 0.226 & \textbf{0.173} & \textbf{0.223}\\
    & 192 & 0.619 & 0.560 & 0.273 & 0.297 & 0.699 & 0.604 & 0.480 & 0.464 & 0.256 & 0.295 & \textbf{0.245} & \textbf{0.285}\\
    & 336 & 0.689 & 0.594 & 0.333 & \textbf{0.325} & 0.773 & 0.620 & 0.581 & 0.519 & 0.338 & 0.351 & \textbf{0.321} & 0.338\\
    & 720 & 0.926 & 0.710 & 0.436 & 0.420 & 1.008 & 0.718 & 0.795 & 0.642 & 0.417 & 0.412 & \textbf{0.414} & \textbf{0.410}\\
    \bottomrule
  \end{tabular}
  \end{small}
  \vspace{-12pt}
\end{table}

\section{Conclusion}
\vspace{-5pt}
This paper addresses time series forecasting from the view of stationarity. Unlike previous studies that simply attenuate non-stationarity leading to over-stationarization, we propose an efficient way to increase series stationarity and renovate the internal mechanism to re-incorporate non-stationary information, thus boosting data predictability and model predictive capability simultaneously. Experimentally, our method shows great generality and performance on six real-world benchmarks. And detailed derivations and ablations are provided to testify the effectiveness of each component in our proposed Non-stationary Transformers framework. In the future, we will explore a more model-agnostic solution for the over-stationarization problem.

\vspace{-5pt}
\section*{Acknowledgments}
\vspace{-5pt}
This work was supported by the National Key Research and Development Plan (2021YFC3000905), National Natural Science Foundation of China (62022050 and 62021002), Beijing Nova Program (Z201100006820041), and BNRist Innovation Fund (BNR2021RC01002).

\clearpage

\small
\bibliographystyle{plain}
\bibliography{ref}

\newpage


\newpage

\appendix

\section{Proof of De-stationary Attention}\label{app:proof}

\begin{definition}
Self-Attention~\cite{NIPS2017_3f5ee243} is defined as:
\begin{equation}
\label{equ:attns}
    \operatorname{Attn}(\mathbf{Q}, \mathbf{K}, \mathbf{V} )=\operatorname{Softmax}\left(\frac{\mathbf{Q} \mathbf{K}^\top}{\sqrt{d_{k}}}\right) \mathbf{V},
\end{equation}
where $\mathbf{Q},\mathbf{K}$ and $\mathbf{V}\in\mathbb{R}^{S\times d_k}$ are length-$S$ query, key and value, where $S$ is the length of input sequence and $d_k$ is the feature dimension, and $\operatorname{Softmax}(\cdot)$ is conducted on each row.
\end{definition}
\vspace{7pt}
\begin{assumption}
\emph{The embedding layer and feed forward layer are functions conducted separately at each time point of the input and hold the linear property.}\label{asm:linear}

For example, the query $\mathbf{Q}$ as the input of the first $\operatorname{Attn(\cdot)}$ layer is obtained by feeding the input $\mathbf{x}=[x_1,x_2,\cdots,x_S]^\top \in \mathbb{R}^{S\times C}$ into the embedding layer $f: \mathbb{R}^{C \times 1} \to \mathbb{R}^{d_k \times 1}$, where $C$ is the number of series variables. And each of the query token in $\mathbf{Q} = [q_1, q_2, ..., q_S]^\top$ can be calculated as $q_i=f(x_i)$ w.r.t. each time point in  $\mathbf{x}=[x_1,x_2,\cdots,x_S]^\top$. Function $f$ holds the linear property means that $f(ax+by)=af(x)+bf(y)$, where $a, b$ are scalars and $x,y$ are vectors.
\end{assumption}
\vspace{5pt}
\begin{assumption}
\emph{Each variable of the input series has the same variance.}\label{asm:share}

For each input time series $\mathbf{x}$, we calculate its mean and variance as follows:
\begin{equation*}
    \mu_{\mathbf{x}} = \frac{1}{S}\sum_{i=1}^S x_i,\ \sigma_{\mathbf{x}}^2 =\frac{1}{S} \sum_{i=1}^S(x_i-\mu_{\mathbf{x}})^2,
\end{equation*}
where $\mu_{\mathbf{x}}, \sigma_{\mathbf{x}} \in \mathbb{R}^{C \times 1}$ is the mean and standard deviation of all $x_i$s. Since it is a convention to conduct normalization on each series variable to avoid certain variable that dominates the scale, we can assume that each variable shares the same variance, and thus $\sigma_{\mathbf{x}}$ is reduced to a scalar.
\end{assumption}
\vspace{7pt}
\begin{theorem}
\begin{equation}\label{equ:insights}
     \operatorname{Softmax}\left( \frac{\mathbf{Q} \mathbf{K}^{\top}}{\sqrt{d_k}}\right) = \operatorname{Softmax}\left( \frac{ \sigma_{\mathbf{x}}^2 \  \mathbf{Q^\prime}\mathbf{K^\prime}^{\top} + \mathbf{1} \mu_{\mathbf{Q}}^{\top} \mathbf{K}^{\top}}{\sqrt{d_k}}\right).
\end{equation}
Equation~\ref{equ:insights} means the $\operatorname{Softmax}\left( \mathbf{Q} \mathbf{K}^{\top} / \sqrt{d_k} \right)$ learned from raw series $\mathbf{x}$ can be calculated by current $\mathbf{Q}^\prime,\mathbf{K}^\prime$ learned from stationarized series $\mathbf{x^{\prime}}$, and the calculation also requires the non-stationary information $\sigma_{\mathbf{x}},\mu_{\mathbf{Q}},\mathbf{K}$ that are eliminated during stationarization.
\end{theorem}

\paragraph{Proof 1. (First layer analysis)} After our stationarization, the model receives the normalized input $\mathbf{x^\prime}=[x^\prime_1,x^\prime_2,...,x^\prime_S]^{\top}$ and each $x^\prime_i = (1/\sigma_{\mathbf{x}}) \odot (x_i-\mu_{\mathbf{x}})$. Based on Assumption~\ref{asm:share}, $\sigma_{\mathbf{x}}$ is reduced to a scalar and we can simplify the normalized input of each time point to $x^\prime_i = (x_i-\mu_{\mathbf{x}})/\sigma_{\mathbf{x}}$. Then $\mathbf{x}^\prime$ is fed into the embedding layer $f$. Based on Assumption~\ref{asm:linear}, we get current query $\mathbf{Q}^{\prime}= [q_1^\prime, ..., q_S^\prime]^\top$ of the first $\operatorname{Attn(\cdot)}$ layer:
$$q_i^\prime
=f(\frac{x_i-\mu_{\mathbf{x}}}{\sigma_{\mathbf{x}}})
=\frac{f(x_i)-f(\mu_{\mathbf{x}})}{\sigma_{\mathbf{x}}}
=\frac{q_i-f(\frac{1}{S}\sum_{i=1}^S x_i)}{\sigma_{\mathbf{x}}}
=\frac{q_i-\frac{1}{S}\sum_{i=1}^Sf(x_i)}{\sigma_{\mathbf{x}}}
=\frac{q_i-\mu_{\mathbf{Q}}}{\sigma_{\mathbf{x}}},$$
where $\mu_{\mathbf{Q}}=\frac{1}{S}\sum_{i=1}^S q_i \in\mathbb{R}^{d_k \times 1}$. Then $\mathbf{Q}^{\prime}= [q_1^\prime, ..., q_S^\prime]^\top$ can be written as $(\mathbf{Q}- \mathbf{1}\mu_{\mathbf{Q}}^\top)/\sigma_{\mathbf{x}}$ and $\mathbf{1} \in \mathbb{R}^{S\times 1}$ is an all-ones vector. And so is the corresponding transformed $\mathbf{K}^{\prime}$. Without the stationarization, the input of $\operatorname{Softmax}(\cdot)$ in Self-Attention should be $(\mathbf{Q}\mathbf{K}^{\top}/\sqrt{d_k})$, while now the attention is calculated based on $\mathbf{Q}^{\prime}, \mathbf{K}^{\prime}$. And we have the following equations:
\begin{equation*}
\begin{split}
    \mathbf{Q}^{\prime}\mathbf{K}^{\prime\top} &=\frac{1}{\sigma_{\mathbf{x}}^2} \left( \mathbf{Q}\mathbf{K}^{\top}-
    \mathbf{1} (\mu_{\mathbf{Q}}^{\top} \mathbf{K}^{\top}) -
    (\mathbf{Q} \mu_{\mathbf{K}}) \mathbf{1}^{\top} + 
    \mathbf{1} (\mu_{\mathbf{Q}}^{\top} \mu_{\mathbf{K}}) \mathbf{1}^{\top} \right), \\
    \operatorname{Softmax}\left( \frac{\mathbf{Q} \mathbf{K}^{\top}}{\sqrt{d_k}}\right) &= 
    \operatorname{Softmax}\left( \frac{
    \sigma_{\mathbf{x}}^2 \  \mathbf{Q^\prime}\mathbf{K^\prime}^{\top} + 
    \mathbf{1} (\mu_{\mathbf{Q}}^{\top} \mathbf{K}^{\top}) +
    (\mathbf{Q} \mu_{\mathbf{K}}) \mathbf{1}^{\top} -
    \mathbf{1} (\mu_{\mathbf{Q}}^{\top} \mu_{\mathbf{K}}) \mathbf{1}^{\top}}{\sqrt{d_k}}
    \right).
\end{split}
\end{equation*}
 We find that $\mathbf{Q} \mu_{\mathbf{K}} \in \mathbb{R}^{S\times 1}$ and $\mu_{\mathbf{Q}}^{\top} \mu_{\mathbf{K}}\in \mathbb{R}$, and they are repeatedly operated on each column and element of $\sigma_{\mathbf{x}}^2\mathbf{Q^\prime}\mathbf{K^\prime}^{\top}\in \mathbb{R}^{S\times S}$. Since $\operatorname{Softmax}(\cdot)$ is invariant to the same translation on the row dimension of input, we have the following equation: 
\begin{equation*}
     \operatorname{Softmax}\left( \frac{\mathbf{Q} \mathbf{K}^{\top}}{\sqrt{d_k}}\right)  = \operatorname{Softmax}\left( \frac{ \sigma_{\mathbf{x}}^2 \  \mathbf{Q^\prime}\mathbf{K^\prime}^{\top} + \mathbf{1} \mu_{\mathbf{Q}}^{\top} \mathbf{K}^{\top}}{\sqrt{d_k}}\right).
\end{equation*}

\paragraph{Proof 2. (Multiple layers analysis)}
We have deduced an equivalent expression of the output of the first $\operatorname{Softmax}(\cdot)$. If we can successfully approximate the attention map that related to $\mathbf{Q}$ and $\mathbf{K}$, we only need to consider $\operatorname{Attn}(\cdot)$ with respect to the change of $\mathbf{V}$. Fortunately, $\operatorname{Attn}(\cdot)$ as the function of $\mathbf{V}$ gives each time point of the output $\mathbf{E}=[e_1,...,e_S]^\top \in \mathbb{R}^{S \times d_k}$ as a simplex:
$$e_j=\left\{ \sum_{i=1}^S w_i v_i |\mathbf{V}=[v_1, v_2, ..., v_S]^\top, \sum_{i=1}^S w_i=1, w_i \ge 0 \right\},$$
which also holds the linear property $f(a\mathbf{V_1}+b\mathbf{V_2})=af(\mathbf{V_1})+bf(\mathbf{V_2})$. Therefore, the $\operatorname{Attn}(\cdot)$ layer is also a function that satisfies our Assumption~\ref{asm:linear}. We will have each time point of the output $\mathbf{E}$ varies linearly with each time point of the input $\mathbf{x}$, and then $\mathbf{E}$ will become the next block's input. As the feed forward layer, residual adding and $\operatorname{Attn}(\cdot)$ layer are the repeating building blocks of Transformer, they also compose a function with linear property as stated in Assumption~\ref{asm:linear}. By the first layer analysis and induction on each layer, Equation~\ref{equ:insights} will holds for $\operatorname{Softmax}(\cdot)$ of all layers under our assumptions. 

\paragraph{Attention design} Based on the analysis, we develop De-stationary Attention as:
\begin{equation}\label{equ:nsattns}
  \begin{split}
    & \log{\tau} = \operatorname{MLP}(\sigma_\mathbf{x}, \mathbf{x}), \mathbf{\Delta} = \operatorname{MLP}(\mu_\mathbf{x}, \mathbf{x}), \\
    \operatorname{Attn} (\mathbf{Q^\prime}, & \mathbf{K^\prime}, \mathbf{V^\prime}, \tau, \mathbf{\Delta} ) = \operatorname{Softmax}\left(\frac{\tau \ \mathbf{Q^\prime} \mathbf{K^\prime}^\top+ \mathbf{1}\mathbf{\Delta}^{\top}}{\sqrt{d_{k}}}\right) \mathbf{V^\prime},
  \end{split}
\end{equation}
where $\tau\in\mathbb{R}^{+}$ and $\mathbf{\Delta}\in\mathbb{R}^{S\times 1}$ is defined as the scaling and shifting de-stationary factors respectively to approximate $\sigma_{\mathbf{x}}^2$ and $\mathbf{K}\mu_\mathbf{Q}$ under the real scenario. Since the key to making Equation~\ref{equ:insights} established is to approximate the attention map successfully, we apply a direct deep learning implementation. To be concisely, we use a multi-layer perceptron as the projector to learn de-stationary factors $\tau, \mathbf{\Delta}$ from the statistics $\mu_{\mathbf{x}}, \sigma_{\mathbf{x}}$ and unstationarized $\mathbf{x}$. De-stationary Attention learns the temporal dependencies from both stationarized series $\mathbf{Q}'$, $\mathbf{K'}$ and non-stationary series $\mathbf{x}$, $\mu_{\mathbf{x}}$, $\sigma_{\mathbf{x}}$, and multiplies by the stationarized values $\mathbf{V^\prime}$ to keep the linear property. It can benefit from the predictability of stationarized series and re-incorporate the inherent non-stationarity of raw series simultaneously.

\section{Hyperparameter Sensitivity}\label{app:hyp}
We verify the robustness of the proposed Non-stationary Transformers framework with respect to hyper-parameter $dim$, which is the hidden layer dimension of the MLP projector that learns de-stationary factors. Considering the efficiency of hyperparameters search, we fix the number of hidden layers, and the hidden layer dimension varies in $\{64, 128, 256\}$. The results are shown in Table~\ref{tab:different_dim}. For datasets with relatively high non-stationarity (Exchange and ILI), large $dim$ would be a better choice, which indicates that non-stationary information entangled with unstationarized input should be learned by a projector with big capacity. Besides, as the dataset presents higher non-stationarity, the influence of de-stationary project design becomes more significant.
\begin{table}[htbp]
  \caption{The performance of Non-stationary Transformers under different choices of the hidden layer dimension  ($dim$) in the projector. We adopt the forecasting setting as input-36-predict-48 for the ILI dataset and input-96-predict-336 for the other datasets.}\label{tab:different_dim}
  \centering
  \begin{small}
  \renewcommand{\multirowsetup}{\centering}
  \setlength{\tabcolsep}{4.0pt}
  \begin{tabular}{c|cc|cc|cc|cc|cc|cccc}
    \toprule
    Dataset & \multicolumn{2}{c}{Exchange} & \multicolumn{2}{c}{ILI} & \multicolumn{2}{c}{ETTm2}  & \multicolumn{2}{c}{Electricity}  & \multicolumn{2}{c}{Traffic} & \multicolumn{2}{c}{Weather}   \\
    \cmidrule(lr){2-3} \cmidrule(lr){4-5}\cmidrule(lr){6-7} \cmidrule(lr){8-9}\cmidrule(lr){10-11}\cmidrule(lr){12-13}
    Metric & MSE & MAE & MSE & MAE & MSE & MAE & MSE & MAE & MSE & MAE & MSE & MAE  \\
    \toprule
    $dim=\ \ 64$ & 0.448 & 0.493 & 2.067 & 0.908 &  \textbf{0.334} & \textbf{0.361} & \textbf{0.200} & \textbf{0.304}  & 0.629 & 0.345  & \textbf{0.321} & \textbf{0.338}  \\
    $dim=128$    & 0.432 & 0.477 & \textbf{2.010} & \textbf{0.900} &  0.370 & 0.388 & 0.201 & 0.301  & \textbf{0.618} & \textbf{0.328}  & 0.340 & 0.354 \\
    $dim=256$    & \textbf{0.421} & \textbf{0.476} & 2.223 & 0.928 &  0.367 & 0.381 & 0.201 & 0.304  & 0.631 & 0.351  & 0.333 & 0.347 \\
    \bottomrule
  \end{tabular}
  \end{small}
  \vspace{-15pt}
\end{table}

\section{Supplementary of Main Results}

\subsection{Multivariable Forecasting Results}\label{app:add}
As shown in Table~\ref{tab:full_results}, we list additional benchmark on the ETT datasets~\cite{haoyietal-informer-2021}, which includes the hourly recorded ETTh1/ETTh2 and 15-minutely recorded ETTm1. Non-stationary Transformer also achieves remarkable improvement over the state-of-the-art on various forecasting horizons. For the input-96-predict-336 long-term setting, Non-stationary Transformer surpasses previous best results by \textbf{4.4\%}($0.615\to 0.588$) in ETTh1, \textbf{3.5\%}($0.572\to 0.552$) in ETTh2 and \textbf{26.7\%}($0.675\to 0.495$) MSE reduction in ETTm1. The overall results show averaged \textbf{11.5\%} MSE reduction over previous state-of-the-art deep forecasting models.

We also list additional model comparison in Table~\ref{tab:additional_baseline}, including the concurrent work FEDformer~\cite{zhou2022fedformer}, and non-Transformer models LSSL~\cite{gu2021combining} and GRU~\cite{cho2014learning}. Our method still outperforms these models in most cases (83\%). Notably, LSSL~\cite{gu2021combining} achieves good performance on Weather~\cite{weatherdata} dataset with the highest stationarity but poorly performs on others, especially non-stationary datasets.

\begin{table}[htbp]
  \caption{Forecasting results comparison under different prediction lengths $O \in \{96,192,336,720\}$ on ETT dataset. The input sequence length is set to 96.}\label{tab:full_results}
  \centering
  \begin{threeparttable}
  \begin{small}
  \renewcommand{\multirowsetup}{\centering}
  \setlength{\tabcolsep}{2.6pt}
  \begin{tabular}{c|c|cccccccccccccc}
    \toprule
    \multicolumn{2}{c}{Models} & \multicolumn{2}{c}{\textbf{Ours}} & \multicolumn{2}{c}{\scalebox{0.8}{Autoformer\cite{wu2021autoformer}}} & \multicolumn{2}{c}{\scalebox{0.8}{Pyraformer\cite{liu2021pyraformer}}} &  \multicolumn{2}{c}{Informer\cite{haoyietal-informer-2021}} & \multicolumn{2}{c}{LogTrans\cite{2019Enhancing}}  & \multicolumn{2}{c}{Reformer\cite{kitaev2020reformer}} & \multicolumn{2}{c}{LSTNet\cite{2018Modeling}}  \\
    \cmidrule(lr){3-4} \cmidrule(lr){5-6}\cmidrule(lr){7-8} \cmidrule(lr){9-10}\cmidrule(lr){11-12}\cmidrule(lr){13-14}\cmidrule(lr){15-16}
    \multicolumn{2}{c}{Metric} & MSE & MAE & MSE & MAE & MSE & MAE & MSE & MAE & MSE & MAE & MSE & MAE & MSE & MAE  \\
    \toprule
    \multirow{4}{*}{\rotatebox{90}{ETTh1}}
    &  96 & \textbf{0.513} & \textbf{0.491} & 0.536 & 0.548 & 0.783 & 0.657 & 0.984 & 0.786 & 0.767 & 0.758 & 0.773 & 0.640 & 1.457 & 0.961\\
    & 192 & \textbf{0.534} & \textbf{0.504} & 0.543 & 0.551 & 0.863 & 0.709 & 1.027 & 0.791 & 1.003 & 0.849 & 0.910 & 0.704 & 1.998 & 1.215\\
    & 336 & \textbf{0.588} & \textbf{0.535} & 0.615 & 0.592 & 0.941 & 0.753 & 1.032 & 0.774 & 1.362 & 0.952 & 1.000 & 0.760 & 2.655 & 1.369\\
    & 720 & 0.643 & 0.616 & \textbf{0.599} & \textbf{0.600} & 1.042 & 0.819 & 1.169 & 0.858 & 1.397 & 1.291 & 1.242 & 0.860 & 2.143 & 1.380\\
    \midrule
    \multirow{4}{*}{\rotatebox{90}{ETTh2}}  
    & 96    & \textbf{0.476} & \textbf{0.458} & 0.492 & 0.517 & 1.380 & 0.943 & 2.826 & 1.330 & 0.829 & 0.751 & 1.595 & 1.031 & 3.568 & 1.688\\
    & 192   & \textbf{0.512} & \textbf{0.493} & 0.556 & 0.551 & 3.809 & 1.634 & 6.186 & 2.070 & 1.807 & 1.036 & 2.671 & 1.300 & 3.243 & 2.514\\
    & 336   & \textbf{0.552} & \textbf{0.551} & 0.572 & 0.578 & 4.282 & 1.792 & 5.268 & 1.942 & 3.875 & 1.763 & 2.596 & 1.297 & 2.544 & 2.591\\
    & 720   & \textbf{0.562} & \textbf{0.560} & 0.580 & 0.588 & 4.252 & 1.790 & 3.667 & 1.616 & 3.913 & 1.552 & 2.647 & 1.304 & 4.625 & 3.709\\
    \midrule
    \multirow{4}{*}{\rotatebox{90}{ETTm1}} 
    &  96 & \textbf{0.386} & \textbf{0.398} & 0.523 & 0.488 & 0.536 & 0.506 & 0.615 & 0.556 & 0.588 & 0.593 & 0.778 & 0.623 & 2.003 & 1.218\\
    & 192 & \textbf{0.459} & \textbf{0.444} & 0.543 & 0.498 & 0.539 & 0.520 & 0.723 & 0.620 & 0.769 & 0.793 & 0.929 & 0.707 & 2.764 & 1.544\\
    & 336 & \textbf{0.495} & \textbf{0.464} & 0.675 & 0.551 & 0.720 & 0.635 & 1.300 & 0.908 & 1.462 & 1.320 & 1.016 & 0.733 & 1.257 & 2.076\\
    & 720 & \textbf{0.585} & \textbf{0.516} & 0.720 & 0.573 & 0.940 & 0.740 & 0.972 & 0.744 & 1.669 & 1.461 & 1.122 & 0.793 & 1.917 & 2.941\\
    \bottomrule
  \end{tabular}
  \end{small}
  \end{threeparttable}
\end{table}

\begin{table}[htbp]
  \caption{Forecasting results comparison with additional baseline forecasting models.}\label{tab:additional_baseline}
  \centering
  \begin{small}
  \renewcommand{\multirowsetup}{\centering}
  \setlength{\tabcolsep}{10.5pt}
  \begin{tabular}{c|c|cc|cc|cc|cc}
    \toprule
    \multicolumn{2}{c}{Models} &
    \multicolumn{2}{c}{Ours} &
    \multicolumn{2}{c}{FEDformer~\cite{zhou2022fedformer}} & 
    \multicolumn{2}{c}{LSSL~\cite{gu2021combining}} &
    \multicolumn{2}{c}{GRU~\cite{cho2014learning}} \\
    \cmidrule(lr){3-4} 
    \cmidrule(lr){5-6}
    \cmidrule(lr){7-8} 
    \cmidrule(lr){9-10} 
    \multicolumn{2}{c}{Metric} & MSE & MAE & MSE & MAE & MSE & MAE & MSE & MAE\\
    \toprule
    \multirow{4}{*}{\rotatebox{90}{Exchange}} 
    & 96  &  \textbf{0.111} & \textbf{0.237} & 0.148 &  0.271 & 0.395 &  0.474 & 1.453 &  1.049\\
    & 192 &  \textbf{0.219} & \textbf{0.335} & 0.271 &  0.380 & 0.776 &  0.698 & 1.846 &  1.179\\
    & 336 &  \textbf{0.421} & \textbf{0.476} & 0.460 & 0.500 & 1.029 &  0.797 & 2.136 &  1.231\\
    & 720 &  \textbf{1.092} & \textbf{0.769} & 1.195 & 0.841 & 2.283 &  1.222 & 2.984 &  1.427\\
    \midrule
    \multirow{4}{*}{\rotatebox{90}{ILI}} 
    & 24 & \textbf{2.294} & \textbf{0.945} & 3.228 &  1.260 & 4.381 &  1.425 & 5.914 &  1.734\\
    & 36 & \textbf{1.825} & \textbf{0.848} & 2.679 &  1.080 & 4.442 &  1.416 & 6.631 &  1.845\\
    & 48 & \textbf{2.010} & \textbf{0.900} & 2.622 &  1.078 & 4.559 &  1.443 & 6.736 &  1.857	\\
    & 60 & \textbf{2.178} & \textbf{0.963} & 2.857 &  1.157 & 4.651 &  1.474& 6.870 &  1.879\\
    \midrule
    \multirow{4}{*}{\rotatebox{90}{ETTm2}} 
    & 96 &  \textbf{0.192} & \textbf{0.274} & 0.203 &  0.287 & 0.243 &  0.342 & 2.041 &  1.073	\\
    & 192 & 0.280 & 0.339 & \textbf{0.269} &  \textbf{0.328} & 0.392 &  0.448 & 2.249 &  1.112\\
    & 336 & 0.334 & \textbf{0.361} & \textbf{0.325} &  0.366 & 0.932 &  0.724 & 2.568 &  1.238	\\
    & 720 & \textbf{0.417} & \textbf{0.413} & 0.421 &  0.415 & 1.372 &  0.879 & 2.720 &  1.287\\
    \midrule
    \multirow{4}{*}{\rotatebox{90}{Electricity}} 
    & 96  &  \textbf{0.169} & \textbf{0.273} & 0.193 &  0.308 & 0.300 &  0.392 & 0.375 &  0.437		\\
    & 192 &  \textbf{0.182} & \textbf{0.286} & 0.201 &  0.315 & 0.297 &  0.390 & 0.442 &  0.473	\\
    & 336 &  \textbf{0.200} & \textbf{0.304} & 0.214 &  0.329 & 0.317 &  0.403 & 0.439 &  0.473\\
    & 720 &  \textbf{0.222} & \textbf{0.321} & 0.246 &  0.355 & 0.338 &  0.417 & 0.980 &  0.814\\
    \midrule
    \multirow{4}{*}{\rotatebox{90}{Traffic}} 
    & 96  &  0.612 & \textbf{0.338} & \textbf{0.587} &  0.366 & 0.798 &  0.436& 0.843 &  0.453\\
    & 192 &  0.613 & \textbf{0.340} & \textbf{0.604} &  0.373&  0.849 &  0.481 & 0.847 &  0.453\\
    & 336 &  \textbf{0.618} & \textbf{0.328} & 0.621 &  0.383 & 0.828 &  0.476 & 0.853 &  0.455\\
    & 720 &  0.653 & \textbf{0.355} & \textbf{0.626} &  0.382 & 0.854 &  0.489 & 1.500 &  0.805\\
    \midrule
    \multirow{4}{*}{\rotatebox{90}{Weather}} 
    & 96 &  \textbf{0.173} & \textbf{0.223} & 0.217 &  0.296 & 0.174 &  0.252 & 0.369 &  0.406\\
    & 192 & 0.245 & \textbf{0.285} & 0.276 &  0.336 & \textbf{0.238} &  0.313 & 0.416 &  0.435\\
    & 336 & 0.321 & \textbf{0.338} & 0.339 &  0.380 & \textbf{0.287} &  0.355 & 0.455 &  0.454\\
    & 720 & 0.414 & \textbf{0.410} & 0.403 &  0.428 & \textbf{0.384} &  0.415 & 0.535 & 0.520\\
    \bottomrule
  \end{tabular}
  \end{small}
\end{table}

\subsection{Performance of Non-stationary Transformer and Variants}\label{app:full}
We apply our proposed Non-stationary Transformers framework to six Transformer variants: Transformer~\cite{NIPS2017_3f5ee243}, Informer~\cite{haoyietal-informer-2021}, Reformer~\cite{kitaev2020reformer}, Autoformer~\cite{wu2021autoformer},
ETSformer~\cite{woo2022etsformer} and FEDformer~\cite{zhou2022fedformer}. The averaged results are shown in Table~\ref{tab:boost} due to the limited pages. We provide supplementary forecasting results in Table~\ref{tab:full_boost} and Table~\ref{tab:additional_boost}. The experimental results demonstrate that our Non-stationary Transformers framework can consistently promotes these Transformer variants, even on the concurrent work ETSformer and FEDformer.

\begin{table}[htbp]
  \caption{Detailed forecasting performance of Non-stationary Transformers. We report the MSE/MAE of different prediction lengths $O \in \{96,192,336,720\}$ and $\{24,36,48,60\}$ for comparison. The input sequence length is set to 36 for ILI and 96 for the others.}\label{tab:full_boost}
  \centering
  \begin{small}
  \renewcommand{\multirowsetup}{\centering}
  \setlength{\tabcolsep}{1.8pt}
  \begin{tabular}{c|c|cc|cc|cc|cc}
    \toprule
    \multicolumn{2}{c}{Models} &
    \multicolumn{2}{c}{Transformer + Ours } &
    \multicolumn{2}{c}{Informer + Ours} &
    \multicolumn{2}{c}{Reformer + Ours} & 
    \multicolumn{2}{c}{Autoformer + Ours} \\
    \cmidrule(lr){3-4} \cmidrule(lr){5-6}
    \cmidrule(lr){7-8} \cmidrule(lr){9-10}
    \multicolumn{2}{c}{Metric} & MSE & MAE & MSE & MAE & MSE & MAE & MSE & MAE  \\
    \toprule
    \multirow{4}{*}{\rotatebox{90}{Exchange}} 
    & 96 &  0.111\scalebox{0.8}{$\pm$0.015} & 0.237\scalebox{0.8}{$\pm$0.010} & 0.129\scalebox{0.8}{$\pm$0.018} & 0.258\scalebox{0.8}{$\pm$0.012} & 0.128\scalebox{0.8}{$\pm$0.019} & 0.258\scalebox{0.8}{$\pm$0.012} & 0.171\scalebox{0.8}{$\pm$0.005} & 0.276\scalebox{0.8}{$\pm$0.006} \\
    & 192 & 0.219\scalebox{0.8}{$\pm$0.031} & 0.335\scalebox{0.8}{$\pm$0.020} & 0.251\scalebox{0.8}{$\pm$0.042} & 0.354\scalebox{0.8}{$\pm$0.035} & 0.246\scalebox{0.8}{$\pm$0.045} & 0.356\scalebox{0.8}{$\pm$0.037} & 0.273\scalebox{0.8}{$\pm$0.005} & 0.365\scalebox{0.8}{$\pm$0.007} \\
    & 336 & 0.421\scalebox{0.8}{$\pm$0.032} & 0.476\scalebox{0.8}{$\pm$0.022} & 0.373\scalebox{0.8}{$\pm$0.047} & 0.434\scalebox{0.8}{$\pm$0.032} & 0.422\scalebox{0.8}{$\pm$0.039} & 0.478\scalebox{0.8}{$\pm$0.030} & 0.481\scalebox{0.8}{$\pm$0.010} & 0.573\scalebox{0.8}{$\pm$0.009} \\
    & 720 & 1.092\scalebox{0.8}{$\pm$0.027} & 0.769\scalebox{0.8}{$\pm$0.024} & 1.229\scalebox{0.8}{$\pm$0.035} & 0.795\scalebox{0.8}{$\pm$0.049} & 1.050\scalebox{0.8}{$\pm$0.050} & 0.781\scalebox{0.8}{$\pm$0.047} & 1.024\scalebox{0.8}{$\pm$0.012} & 0.751\scalebox{0.8}{$\pm$0.012} \\
    \midrule
    \multirow{4}{*}{\rotatebox{90}{ILI}} 
    & 24   & 2.294\scalebox{0.8}{$\pm$0.152} & 0.945\scalebox{0.8}{$\pm$0.041} & 2.856\scalebox{0.8}{$\pm$0.177} & 1.071\scalebox{0.8}{$\pm$0.067}  & 3.206\scalebox{0.8}{$\pm$0.277} & 1.131\scalebox{0.8}{$\pm$0.079} & 3.029\scalebox{0.8}{$\pm$0.116} & 1.166\scalebox{0.8}{$\pm$0.028} \\
    & 36   & 1.825\scalebox{0.8}{$\pm$0.128} & 0.848\scalebox{0.8}{$\pm$0.033} & 1.805\scalebox{0.8}{$\pm$0.143} & 0.860\scalebox{0.8}{$\pm$0.051}  & 2.750\scalebox{0.8}{$\pm$0.161} & 1.018\scalebox{0.8}{$\pm$0.074} & 2.648\scalebox{0.8}{$\pm$0.134} & 1.023\scalebox{0.8}{$\pm$0.032} \\
    & 48   & 2.010\scalebox{0.8}{$\pm$0.134} & 0.900\scalebox{0.8}{$\pm$0.035} & 1.780\scalebox{0.8}{$\pm$0.194} & 0.849\scalebox{0.8}{$\pm$0.054}  & 2.710\scalebox{0.8}{$\pm$0.184} & 1.017\scalebox{0.8}{$\pm$0.050} & 2.202\scalebox{0.8}{$\pm$0.161} & 0.965\scalebox{0.8}{$\pm$0.038} \\
    & 60   & 2.178\scalebox{0.8}{$\pm$0.146} & 0.963\scalebox{0.8}{$\pm$0.037} & 2.058\scalebox{0.8}{$\pm$0.173} & 0.933\scalebox{0.8}{$\pm$0.058}  & 2.792\scalebox{0.8}{$\pm$0.153} & 1.095\scalebox{0.8}{$\pm$0.047} & 2.302\scalebox{0.8}{$\pm$0.088} & 1.003\scalebox{0.8}{$\pm$0.024}\\
    \midrule 
    \multirow{4}{*}{\rotatebox{90}{ETTm2}} 
    & 96  & 0.192\scalebox{0.8}{$\pm$0.023} & 0.274\scalebox{0.8}{$\pm$0.016}  & 0.241\scalebox{0.8}{$\pm$0.035} & 0.312\scalebox{0.8}{$\pm$0.025} & 0.209\scalebox{0.8}{$\pm$0.040} & 0.287\scalebox{0.8}{$\pm$0.028} & 0.236\scalebox{0.8}{$\pm$0.022} & 0.319\scalebox{0.8}{$\pm$0.019} \\
    & 192 & 0.280\scalebox{0.8}{$\pm$0.021} & 0.339\scalebox{0.8}{$\pm$0.013}  & 0.433\scalebox{0.8}{$\pm$0.036} & 0.420\scalebox{0.8}{$\pm$0.025} & 0.435\scalebox{0.8}{$\pm$0.037} & 0.421\scalebox{0.8}{$\pm$0.026} & 0.263\scalebox{0.8}{$\pm$0.026} & 0.316\scalebox{0.8}{$\pm$0.025} \\
    & 336 & 0.334\scalebox{0.8}{$\pm$0.011} & 0.361\scalebox{0.8}{$\pm$0.017}  & 0.507\scalebox{0.8}{$\pm$0.032} & 0.464\scalebox{0.8}{$\pm$0.023} & 0.559\scalebox{0.8}{$\pm$0.033} & 0.475\scalebox{0.8}{$\pm$0.024} & 0.320\scalebox{0.8}{$\pm$0.019} & 0.349\scalebox{0.8}{$\pm$0.014} \\
    & 720 & 0.417\scalebox{0.8}{$\pm$0.009} & 0.413\scalebox{0.8}{$\pm$0.011}  & 0.659\scalebox{0.8}{$\pm$0.019} & 0.539\scalebox{0.8}{$\pm$0.028} & 0.769\scalebox{0.8}{$\pm$0.021} & 0.582\scalebox{0.8}{$\pm$0.021} & 0.402\scalebox{0.8}{$\pm$0.015} & 0.396\scalebox{0.8}{$\pm$0.010} \\
    \midrule
    \multirow{4}{*}{\rotatebox{90}{Electricity}} 
    & 96  & 0.169\scalebox{0.8}{$\pm$0.008} & 0.273\scalebox{0.8}{$\pm$0.002} & 0.195\scalebox{0.8}{$\pm$0.008} & 0.302\scalebox{0.8}{$\pm$0.003} & 0.190\scalebox{0.8}{$\pm$0.007} & 0.293\scalebox{0.8}{$\pm$0.004} & 0.193\scalebox{0.8}{$\pm$0.009} & 0.295\scalebox{0.8}{$\pm$0.003} \\
    & 192 & 0.182\scalebox{0.8}{$\pm$0.007} & 0.286\scalebox{0.8}{$\pm$0.003} & 0.215\scalebox{0.8}{$\pm$0.007} & 0.321\scalebox{0.8}{$\pm$0.006} & 0.199\scalebox{0.8}{$\pm$0.009} & 0.301\scalebox{0.8}{$\pm$0.008} & 0.211\scalebox{0.8}{$\pm$0.006} & 0.310\scalebox{0.8}{$\pm$0.007} \\
    & 336 & 0.200\scalebox{0.8}{$\pm$0.005} & 0.304\scalebox{0.8}{$\pm$0.005} & 0.235\scalebox{0.8}{$\pm$0.006} & 0.339\scalebox{0.8}{$\pm$0.006} & 0.208\scalebox{0.8}{$\pm$0.005} & 0.310\scalebox{0.8}{$\pm$0.005} & 0.220\scalebox{0.8}{$\pm$0.005} & 0.316\scalebox{0.8}{$\pm$0.004} \\
    & 720 & 0.222\scalebox{0.8}{$\pm$0.016} & 0.321\scalebox{0.8}{$\pm$0.013} & 0.260\scalebox{0.8}{$\pm$0.014} & 0.358\scalebox{0.8}{$\pm$0.014} & 0.226\scalebox{0.8}{$\pm$0.015} & 0.326\scalebox{0.8}{$\pm$0.018} & 0.241\scalebox{0.8}{$\pm$0.019} & 0.337\scalebox{0.8}{$\pm$0.017} \\
    \midrule
    \multirow{4}{*}{\rotatebox{90}{Traffic}} 
    & 96  & 0.612\scalebox{0.8}{$\pm$0.019} & 0.338\scalebox{0.8}{$\pm$0.014} & 0.649\scalebox{0.8}{$\pm$0.028} & 0.370\scalebox{0.8}{$\pm$0.016} & 0.669\scalebox{0.8}{$\pm$0.037} & 0.364\scalebox{0.8}{$\pm$0.020} & 0.604\scalebox{0.8}{$\pm$0.027} & 0.342\scalebox{0.8}{$\pm$0.012} \\
    & 192 & 0.613\scalebox{0.8}{$\pm$0.028} & 0.340\scalebox{0.8}{$\pm$0.018} & 0.689\scalebox{0.8}{$\pm$0.035} & 0.393\scalebox{0.8}{$\pm$0.019} & 0.680\scalebox{0.8}{$\pm$0.036} & 0.369\scalebox{0.8}{$\pm$0.022} & 0.607\scalebox{0.8}{$\pm$0.034} & 0.383\scalebox{0.8}{$\pm$0.020} \\
    & 336 & 0.618\scalebox{0.8}{$\pm$0.018} & 0.328\scalebox{0.8}{$\pm$0.012} & 0.755\scalebox{0.8}{$\pm$0.055} & 0.431\scalebox{0.8}{$\pm$0.054} & 0.688\scalebox{0.8}{$\pm$0.038} & 0.371\scalebox{0.8}{$\pm$0.033} & 0.611\scalebox{0.8}{$\pm$0.019} & 0.353\scalebox{0.8}{$\pm$0.010} \\
    & 720 & 0.653\scalebox{0.8}{$\pm$0.014} & 0.355\scalebox{0.8}{$\pm$0.003} & 0.783\scalebox{0.8}{$\pm$0.026} & 0.440\scalebox{0.8}{$\pm$0.004} & 0.692\scalebox{0.8}{$\pm$0.019} & 0.385\scalebox{0.8}{$\pm$0.014} & 0.653\scalebox{0.8}{$\pm$0.014} & 0.376\scalebox{0.8}{$\pm$0.013} \\
    \midrule
    \multirow{4}{*}{\rotatebox{90}{Weather}}  
    & 96  & 0.173\scalebox{0.8}{$\pm$0.006}  & 0.223\scalebox{0.8}{$\pm$0.004} & 0.186\scalebox{0.8}{$\pm$0.017} & 0.235\scalebox{0.8}{$\pm$0.014} & 0.195\scalebox{0.8}{$\pm$0.020} & 0.242\scalebox{0.8}{$\pm$0.013} & 0.215\scalebox{0.8}{$\pm$0.024} & 0.263\scalebox{0.8}{$\pm$0.019} \\
    & 192 & 0.245\scalebox{0.8}{$\pm$0.014}  & 0.285\scalebox{0.8}{$\pm$0.015} & 0.259\scalebox{0.8}{$\pm$0.024} & 0.292\scalebox{0.8}{$\pm$0.019} & 0.255\scalebox{0.8}{$\pm$0.027} & 0.289\scalebox{0.8}{$\pm$0.023} & 0.257\scalebox{0.8}{$\pm$0.027} & 0.296\scalebox{0.8}{$\pm$0.018} \\
    & 336 & 0.321\scalebox{0.8}{$\pm$0.016}  & 0.338\scalebox{0.8}{$\pm$0.023} & 0.295\scalebox{0.8}{$\pm$0.026} & 0.317\scalebox{0.8}{$\pm$0.018} & 0.306\scalebox{0.8}{$\pm$0.030} & 0.323\scalebox{0.8}{$\pm$0.025} & 0.307\scalebox{0.8}{$\pm$0.009} & 0.321\scalebox{0.8}{$\pm$0.011} \\
    & 720 & 0.414\scalebox{0.8}{$\pm$0.032}  & 0.410\scalebox{0.8}{$\pm$0.031} & 0.361\scalebox{0.8}{$\pm$0.020} & 0.362\scalebox{0.8}{$\pm$0.022} & 0.388\scalebox{0.8}{$\pm$0.024} & 0.376\scalebox{0.8}{$\pm$0.026} & 0.364\scalebox{0.8}{$\pm$0.006} & 0.357\scalebox{0.8}{$\pm$0.007} \\
    \bottomrule
  \end{tabular}
  \vspace{-10pt}
  \end{small}
\end{table}

\begin{table}[tbp]
  \caption{Performance promotion by applying the proposed framework to concurrent ETSformer and FEDformer. We report the averaged MSE/MAE of all prediction lengths (stated in Table~\ref{tab:results}) and the relative MSE reduction ratios (Promotion) by our framework.}
  \label{tab:additional_boost}
  \centering
  \begin{threeparttable}
  \begin{small}
  \renewcommand{\multirowsetup}{\centering}
  \setlength{\tabcolsep}{3.5pt}
  \begin{tabular}{c|cc|cc|cc|cc|cc|cccc}
    \toprule
    Dataset & \multicolumn{2}{c}{Exchange} & \multicolumn{2}{c}{ILI} & \multicolumn{2}{c}{ETTm2}  & \multicolumn{2}{c}{Electricity} & \multicolumn{2}{c}{Traffic} & \multicolumn{2}{c}{Weather}  \\
    \cmidrule(lr){2-3} \cmidrule(lr){4-5}\cmidrule(lr){6-7} \cmidrule(lr){8-9}\cmidrule(lr){10-11}\cmidrule(lr){12-13}
    Model & MSE & MAE & MSE & MAE & MSE & MAE & MSE & MAE & MSE & MAE & MSE & MAE  \\
    \toprule
    ETSformer~\cite{woo2022etsformer} & 0.410 & 0.427 & 2.619 & 1.034&0.293&0.342&0.208&0.323&0.629	&0.403&0.271&0.334\\
    \textbf{ + Ours } & 0.369 & 0.407 & 2.353& 1.017&0.290&0.334&0.203	&0.314&0.618&0.380&0.254&0.293\\
    \cmidrule(lr){1-13}
    Promotion & \multicolumn{2}{c|}{\textbf{10.00\%}} & \multicolumn{2}{c|}{\textbf{10.16\%}} & \multicolumn{2}{c|}{0.77\%} & \multicolumn{2}{c|}{\textbf{2.17\%}} & \multicolumn{2}{c|}{1.75\%} & \multicolumn{2}{c}{\textbf{6.18\%}} \\
    \midrule
    FEDformer~\cite{zhou2022fedformer} & 0.519 & 0.500&2.847&1.144&0.305&0.349&0.214&0.327&0.610	&0.376&0.309&0.360\\
    \textbf{ + Ours } & 0.500 & 0.487&2.728&1.046&0.312&0.346&0.198&	0.300&0.604&0.362&0.268&0.292\\
    \cmidrule(lr){1-13}
    Promotion & \multicolumn{2}{c|}{\textbf{3.66\%}} & \multicolumn{2}{c|}{\textbf{4.18\%}} & \multicolumn{2}{c|}{-2.38\%} & \multicolumn{2}{c|}{\textbf{7.38\%}} & \multicolumn{2}{c|}{0.86\%} & \multicolumn{2}{c}{\textbf{13.36\%}} \\
    \bottomrule
  \end{tabular}
  \end{small}
  \end{threeparttable}
  \vspace{-13pt}
\end{table}

\subsection{Comparison with Stationarization Methods}\label{app:comp}
We provide full comparison among Non-stationary Transformers and two stationarization methods: Revin\cite{kim2022reversible} and Series Stationarization. The averaged results are shown in Table~\ref{tab:comparison} due to the limited pages. As is listed in Table~\ref{tab:full_comparison}, our framework achieves the state-of-the-art performance especially on datasets with high non-stationarity. For Transformer, the proposed method achieves \textbf{25.6\%}($1.467\to 1.092$) MSE reduction on Exchange under the predict-720 settings, \textbf{10.8\%}($2.572\to 2.294$) on ILI under the predict-24 settings, and \textbf{30.3\%}($0.598\to 0.417$) on ETTm2 under the predict-720 settings. As for Reformer, since De-stationary Attention is not directly deduced from the LSH attention~\cite{kitaev2020reformer}, current approximation as stated in Equation~\ref{equ:insights} may not be the optimal solution, but the introducing of De-stationary Attention still achieves relative \textbf{11.6\%}($0.632\to0.559$) promotion on ETTm2 and \textbf{4.4\%}($2.834\to2.770$) on ILI under the predict-336 setting. The comparison demonstrates De-stationary Attention mechanism can further benefit the predictive ability of Transformers.

\begin{table}[htbp]
  \caption{Detailed forecasting results obtained by applying different methods to Transformer and Reformer. We report the MSE/MAE of different prediction lengths for comparison.}\label{tab:full_comparison}
  \centering
  \begin{threeparttable}
  \begin{small}
  \renewcommand{\multirowsetup}{\centering}
  \setlength{\tabcolsep}{4.0pt}
  \begin{tabular}{c|c|cccccc|cccccc}
    \toprule
    \multicolumn{2}{c}{Base Models} &
    \multicolumn{6}{c}{ Transformer } &
    \multicolumn{6}{c}{ Reformer }  \\
    \cmidrule(lr){3-8}  \cmidrule(lr){9-14}
    \multicolumn{2}{c}{\multirow{2}{*}{Methods}} &
    \multicolumn{2}{c}{\multirow{2}{*}{+ RevIN~\cite{kim2022reversible}}} &
    \multicolumn{2}{c}{\scalebox{0.85}{+ Series}} &
    \multicolumn{2}{c}{\multirow{2}{*}{+ \textbf{Ours}}} &
    \multicolumn{2}{c}{\multirow{2}{*}{+ RevIN~\cite{kim2022reversible}}} & 
    \multicolumn{2}{c}{\scalebox{0.85}{+ Series}} & 
    \multicolumn{2}{c}{\multirow{2}{*}{+ \textbf{Ours}}} \\
    
    \multicolumn{2}{c}{ } & \multicolumn{2}{c}{ }  & \multicolumn{2}{c}{\scalebox{0.85}{Stationarization}} & \multicolumn{2}{c}{ } & \multicolumn{2}{c}{ } & \multicolumn{2}{c}{\scalebox{0.85}{Stationarization}} & \multicolumn{2}{c}{ }  \\
    \cmidrule(lr){3-4} \cmidrule(lr){5-6}
    \cmidrule(lr){7-8} \cmidrule(lr){9-10}
    \cmidrule(lr){11-12} \cmidrule(lr){13-14}
    \multicolumn{2}{c}{Metric} & MSE & MAE & MSE & MAE & MSE & MAE & MSE & MAE & MSE & MAE & MSE & MAE \\
    \toprule
    \multirow{4}{*}{\rotatebox{90}{Exchange}} 
    & 96   & 0.136 & 0.258 & 0.136  & 0.258 & \textbf{0.111} & \textbf{0.237}	&0.133 &0.263 &	0.139& 0.265 & \textbf{0.128} & \textbf{0.258}\\
    & 192  & 0.239 & 0.348 & 0.239  & 0.348 & \textbf{0.219} & \textbf{0.335}	& 0.256 &0.363 &	0.257 &0.364& \textbf{0.246} & \textbf{0.356}\\
    & 336  & 0.425 & 0.479 & 0.425  & 0.479 & \textbf{0.421} & \textbf{0.476}	& 0.426 &0.477 &	0.426& 0.477& \textbf{0.422} & \textbf{0.478}\\
    & 720  & 1.467 & 0.862 & 1.475  & 0.865 & \textbf{1.092} & \textbf{0.769}	& 1.059 &0.786 &	1.059 &0.786& \textbf{1.050} & \textbf{0.781}\\
    \midrule
    \multirow{4}{*}{\rotatebox{90}{ILI}} 
    & 24 & 2.572 & 0.980& 2.573 & 0.980 &	 \textbf{2.294}& \textbf{0.945}	& 3.399 &1.170&		3.399 &1.170  &\textbf{3.206} &\textbf{1.131}\\
    & 36 & 1.955 & 0.870& 1.955 & 0.870 &	\textbf{1.825} &\textbf{0.848}  &2.909 &1.049&		2.909 &1.048  &\textbf{2.750}& \textbf{1.018 }\\
    & 48 & 2.056 & 0.902& 2.057 & 0.902 & \textbf{2.010 }&\textbf{0.900}	& 2.834& 1.067 &		2.832 &1.067&\textbf{2.710}& \textbf{1.017}\\
    & 60 & 2.238 & 0.982& 2.238 & 0.982 & \textbf{2.178}& \textbf{0.963}	& 2.954 &1.099 &		2.952 &1.098&\textbf{2.792}& \textbf{1.095}\\
    \midrule
    \multirow{4}{*}{\rotatebox{90}{ETTm2}} 
    & 96 & 0.267 & 0.317& 0.253 & 0.311  & \textbf{0.192}& \textbf{0.274}  		 & 0.211 & 0.295 &	0.212 & 0.297& \textbf{0.209} &\textbf{0.287}\\
    & 192 & 0.456 & 0.405 & 0.453 & 0.404 &\textbf{0.280}& \textbf{0.339}		& 0.478 &0.426 &	0.477 & 0.426 &\textbf{0.435} &\textbf{0.421}\\
    & 336 & 0.528 & 0.455 & 0.546 &0.461 &\textbf{0.334} & \textbf{0.361}		& 0.632 &0.485 &	0.613 & 0.483 &\textbf{0.559} &\textbf{0.475}\\
    & 720 & 0.589 & 0.487 & 0.593 &0.489 &\textbf{0.417} & \textbf{0.413}		& 0.845 & 0.631 &	0.846 & 0.630 &\textbf{0.769} &\textbf{0.582}\\
    \midrule
    \multirow{4}{*}{\rotatebox{90}{Electricity}} 
    & 96  & 0.172 & 0.275 & 0.171 & 0.275 & \textbf{0.169} & \textbf{0.273}& 0.188 & 0.291  & \textbf{0.184} & \textbf{0.289} & 0.190 & 0.293 \\
    & 192 & 0.192 & 0.296 & 0.192 & 0.296 & \textbf{0.182} & \textbf{0.286} & 0.198 & 0.301 & 0.199 & 0.302 & \textbf{0.198} & \textbf{0.301} \\
    & 336 & 0.207 & 0.306 & 0.208 & 0.306 & \textbf{0.200} & \textbf{0.304} & 0.212 & 0.314 & 0.212 & 0.314 & \textbf{0.208} & \textbf{0.310} \\
    & 720 & 0.217 & 0.316 & \textbf{0.216} & \textbf{0.315} & 0.222 & 0.321 & 0.232 & 0.331 & 0.231 & 0.330 & \textbf{0.226} & \textbf{0.326} \\
    \midrule
    \multirow{4}{*}{\rotatebox{90}{Traffic}} 
    & 96  & 0.620 & 0.341 & 0.614 & 0.337 & \textbf{0.612} & \textbf{0.338}& \textbf{0.650} & \textbf{0.364} 	& 0.655 & 0.366 & 0.669 & 0.364 \\
    & 192 & 0.630 & 0.348 & 0.637 & 0.351 & \textbf{0.613} & \textbf{0.340}	& 0.688 & 0.374 & 0.683 & 0.377 & \textbf{0.680} & \textbf{0.369} \\
    & 336 & 0.656 & 0.360 & 0.653 & 0.359 & \textbf{0.634} & \textbf{0.348}	& 0.708 & 0.383 & 0.704 & 0.383 & \textbf{0.688} & \textbf{0.371} \\
    & 720 & 0.666 & 0.360 & 0.661 & 0.360 & \textbf{0.653} & \textbf{0.355}	 & 0.700 & 0.392 & 0.722 & 0.395& \textbf{0.692} & \textbf{0.385} \\
    \midrule
    \multirow{4}{*}{\rotatebox{90}{Weather}} 
    & 96 & 0.175 & 0.225  & 0.175 & 0.225 & \textbf{0.173} & \textbf{0.223}& \textbf{0.189} & \textbf{0.236}	& 0.190 & 0.237  & 0.195 & 0.242 \\
    & 192 & 0.273 & 0.298 & 0.273 & 0.297 & \textbf{0.245} & \textbf{0.285}	& 0.269 & 0.294 & 0.269 & 0.294 & \textbf{0.255} & \textbf{0.289} \\
    & 336 & 0.333 & 0.326 & 0.333 & 0.325 & \textbf{0.321} & \textbf{0.338}	& 0.312 & 0.328 & 0.313 & 0.329 & \textbf{0.306} & \textbf{0.323} \\
    & 720 & 0.424 & 0.415 & 0.436 & 0.420 & \textbf{0.414} & \textbf{0.410}	& 0.395 & 0.376 & 0.395 & 0.376 & \textbf{0.388} & \textbf{0.376} \\
    \bottomrule
  \end{tabular}
  \end{small}
  \end{threeparttable}
\end{table}

\subsection{Prediction Showcases}
We provide supplementary showcases of predictions given by three models: vanilla Transformer, Transformer with Series Stationarization, and Non-stationary Transformer. We plot the last dimension of forecasting results that comes from the \textit{test set} of ETTm1 for qualitative comparison.

\begin{figure*}[htbp]
\begin{center}
	\centerline{\includegraphics[width=\columnwidth]{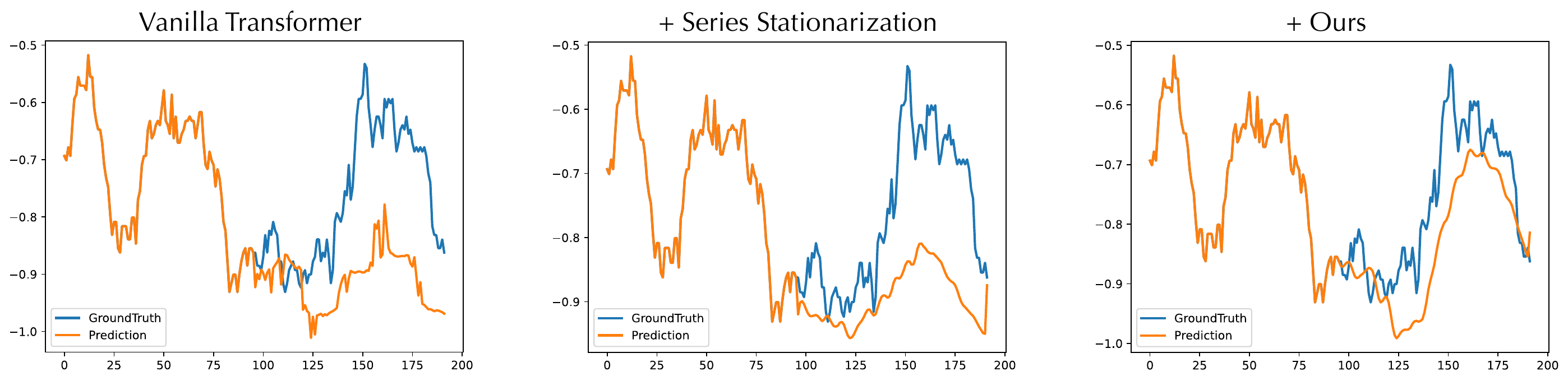}}
	\caption{Visualization of ETTm1 predictions given by different models under the input-96-predict-96 setting. \textcolor{blue}{Blue} lines stand for the ground truth and \textcolor[rgb]{0.8,0.6,0.4}{orange} lines stand for predictions of the model. The first shared part is the time series input with length 96.}
	\label{fig:case96}
\end{center}
\end{figure*}

\begin{figure*}[htbp]
\begin{center}
	\centerline{\includegraphics[width=\columnwidth]{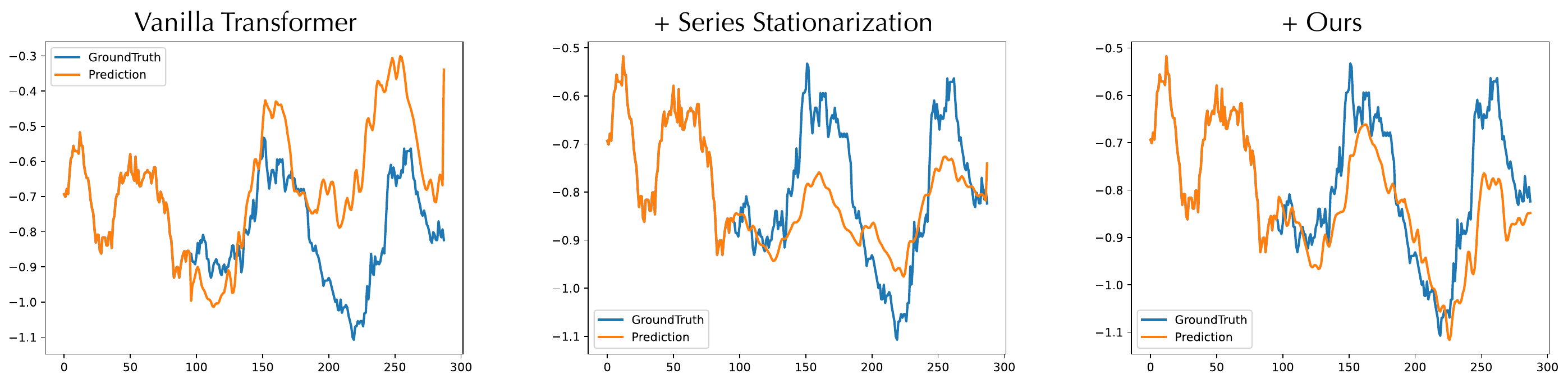}}
	\caption{Visualization of predictions given by models under the input-96-predict-192 setting.}
	\label{fig:case192}
\end{center}
\end{figure*}

\begin{figure*}[htbp]
\begin{center}
	\centerline{\includegraphics[width=\columnwidth]{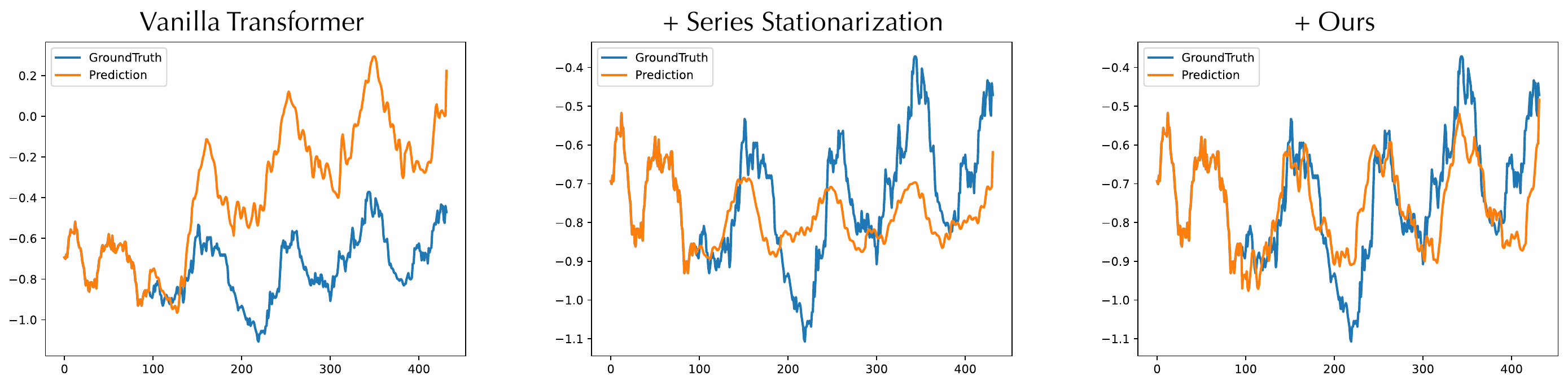}}
	\caption{Visualization of predictions given by models under the input-96-predict-336 setting.}
	\label{fig:case336}
\end{center}
\end{figure*}

\begin{figure*}[htbp]
\begin{center}
	\centerline{\includegraphics[width=\columnwidth]{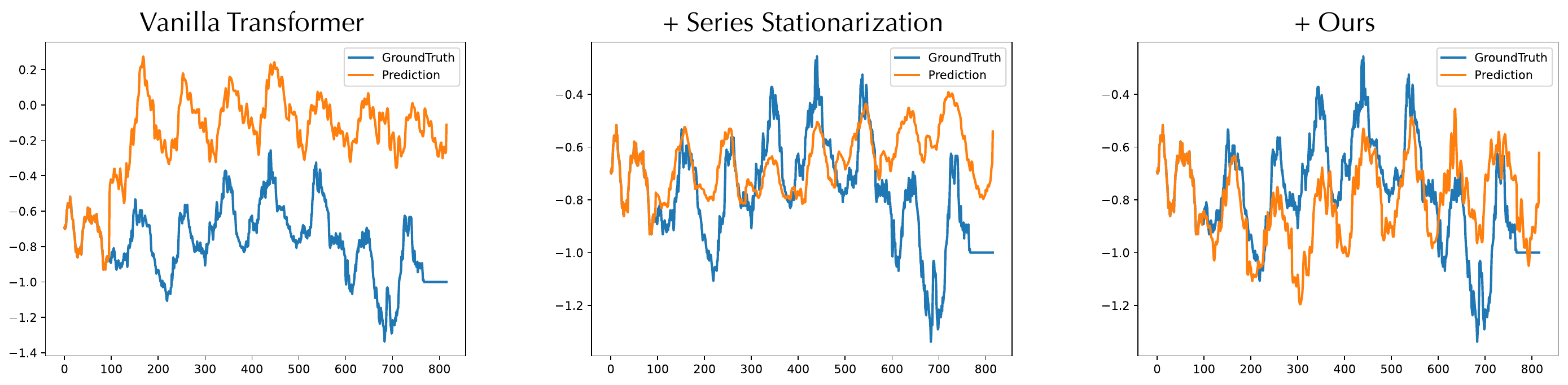}}
	\caption{Visualization of predictions given by models under the input-96-predict-720 setting.}
	\label{fig:case720}
\end{center}
\end{figure*}

As is shown in Figures \ref{fig:case96}, \ref{fig:case192}, \ref{fig:case336}, and \ref{fig:case720}, we find that vanilla Transformer is inclined to output predictions with scale and level far from the ground truth, but its ability to capture local series variation remains strong. While Series Stationarization benefits Transformer by aligning the statistics among each series, the base model neglects the intrinsic non-stationarity of time series and becomes more likely to output stationary but uneventful series. With the help of our framework, the equipped model will be free from the disturbance caused by data non-stationarity and fulfill the potential to capture local variations.


\begin{table}[htbp]
  \caption{Parameters increment and performance promotion of Non-stationary Transformers.}\label{tab:param}
   \vspace{-5pt}
  \centering
  \begin{small}
  \renewcommand{\multirowsetup}{\centering}
  \setlength{\tabcolsep}{6.5pt}
  \begin{tabular}{c|cccccc}
    \toprule
    Models & Transformer & Informer & Reformer & Autoformer & FEDformer & ETSformer \\
    \toprule
    Param increment & \ \ 0.10\% & \ \ 0.09\% & \ \ 0.21\% & \ \ 0.10\% & 0.06\% & 0.19\% \\
    \midrule
    Performance gain & 49.43\% & 47.34\% & 46.89\% & 10.57\% & 4.51\% & 5.17\% \\
    \bottomrule
    \end{tabular}
  \end{small}
\end{table}

\begin{table}[htbp]
  \caption{ADF test statistic of raw series and series processed by our normalization.}
  \label{tab:adf}
  \centering
  \begin{threeparttable}
  \begin{small}
  \renewcommand{\multirowsetup}{\centering}
  \setlength{\tabcolsep}{9pt}
  \begin{tabular}{c|cccccc}
    \toprule
    Dataset & \multicolumn{1}{c}{Exchange} & \multicolumn{1}{c}{ILI} & \multicolumn{1}{c}{ETTm2}  & \multicolumn{1}{c}{Electricity} & \multicolumn{1}{c}{Traffic} & \multicolumn{1}{c}{Weather}  \\
    \toprule
    Raw series & -1.889 & -5.406 & -6.225 & -8.483 & -15.046 & -26.661 \\
    After Normalization &  -9.937 & -10.313 & -33.485 & -20.888 & -18.946 & -35.010 \\
    \bottomrule
  \end{tabular}
  \end{small}
  \end{threeparttable}
\end{table}

\subsection{Efficiency of Non-stationary Transformers}\label{app:eff}
As is shown in Table~\ref{tab:param}, we list the parameters increment and the performance gain (see Promotion in Table~\ref{tab:boost}) of our proposed method. It is obvious that Non-stationary Transformers significantly boosts the forecasting performance by a large margin with hardly any additional parameters.

\section{Ablations}
\subsection{Effects of Series Stationarization}
We propose Series Stationarization, which has no additional learnable parameters, to increase the degree of stationarity and make the time series distribution more stable. As is shown in Table~\ref{tab:adf}, after our normalization module processing, the ADF test statistic of the time series gets obviously smaller, which verifies normalization as an effective design to attenuate the non-stationarity of real world time series.

\subsection{Ablation of De-stationary Factors}
To explore the influence of de-stationary factors, we compare the forecasting results obtained by three variants: only using $\mathbf{\tau}$, only using $\mathbf{\Delta}$, and using both. We conduct experiments on two typical datasets: Exchange (8 variables) and Electricity (321 variables). As is shown in Table~\ref{tab:full_abla}, the forecasting performance will degrade in all cases if we only employ single one of $\tau$ and $\mathbf{\Delta}$, especially without $\tau$ ($0.196\to0.212, 0.441\to0.550$ under the predict-336 setting), which validates the complete form as stated in Equation~\ref{equ:nsattns} is a better choice.

\begin{table}[hbp]
  \caption{Ablation on de-stationary factors: Column (only $\tau$) means only use the scaling de-stationary factor in Equation~\ref{equ:nsattns}, Column (only $\mathbf{\Delta}$) means only use the shifting de-stationary factors, and Column ($\tau$ and $\mathbf{\Delta}$) means use both.}
  \label{tab:full_abla}
  \centering
  \begin{small}
  \renewcommand{\multirowsetup}{\centering}
  \setlength{\tabcolsep}{15.5pt}
  \begin{tabular}{c|c|cc|cc|cc}
    \toprule
    \multicolumn{2}{c}{Models} &
    \multicolumn{2}{c}{Only $\tau$} &
    \multicolumn{2}{c}{Only $\mathbf{\Delta}$} & 
    \multicolumn{2}{c}{$\tau$ and $\mathbf{\Delta}$} \\
    \cmidrule(lr){3-4} 
    \cmidrule(lr){5-6}
    \cmidrule(lr){7-8} 
    \multicolumn{2}{c}{Metric} & MSE & MAE & MSE & MAE & MSE & MAE  \\
    \toprule
    \multirow{4}{*}{\rotatebox{90}{Electricity}} 
    & 96  & 0.177 & 0.279 & 0.186 & 0.287 & \textbf{0.169} & \textbf{0.273} \\
    & 192 & 0.191 & 0.297 & 0.196 & 0.299 & \textbf{0.185} & \textbf{0.289} \\
    & 336 & 0.197 & 0.300 & 0.212 & 0.310 & \textbf{0.196} & \textbf{0.297} \\
    & 720 & 0.221 & 0.320 & 0.227 & 0.326 & \textbf{0.217} & \textbf{0.317} \\
    \midrule
    \multirow{4}{*}{\rotatebox{90}{Exchange}} 
    & 96  & 0.128 & 0.253 & 0.128 & 0.253 & \textbf{0.120} & \textbf{0.247} \\
    & 192 & 0.263 & 0.369 & 0.263 & 0.370 & \textbf{0.250} & \textbf{0.353} \\
    & 336 & 0.446 & 0.491 & 0.550 & 0.553 & \textbf{0.441} & \textbf{0.488} \\
    & 720 & 1.348 & 0.847 & 1.621 & 0.911 & \textbf{1.338} & \textbf{0.847} \\
    \bottomrule
  \end{tabular}
  \end{small}
\end{table}

\section{Non-stationary Transformers: Experimental Details}

\subsection{Detailed Experiment Configurations}\label{app:detail}
We compare each Transformers with and without our framework using the same training strategy. The only hyperparameters for our framework come from the projector design which learns de-stationary factors. We search the hyperparameters as stated in Appendix~\ref{app:hyp}. The best hyperparameter is selected on the \emph{validation set}.

As for other forecast models for the baseline comparison, most of the results are from Autoformer~\cite{wu2021autoformer}. By contacting the authors of Autoformer, we obtain the hyper-parameter selection strategy as follows: for N-BEATS~\cite{oreshkin2019n}, we conduct a grid search for hidden channel in $\{256,512,768\}$, number of layers in $\{2,3,4,5\}$, learning rate in $\{5\times 10^{-5}, 1\times 10^{-4}, 5\times 10^{-4},1\times 10^{-3}\}$. For LSTNet~\cite{2018Modeling}, since the paper also experiments on the Traffic~\cite{trafficdata}, Electricity~\cite{ecldata} and Exchange~\cite{2018Modeling} datasets, the hyper-parameter setting is following the experimental details of the original paper. For N-HiTs~\cite{challu2022n}, ETSformer~\cite{woo2022etsformer}, and FEDformer~\cite{zhou2022fedformer}, as these methods share the same benchmark, we use their official code with three random seeds.

\subsection{Implementation Details of Non-stationary Transformer and Variants} \label{app:impl}
We provide the pseudo-code of Series Stationarization, De-stationary Attention and Non-stationary Transformers in Algorithms~\ref{algo:ssn},~\ref{algo:ssd},~\ref{algo:da} and ~\ref{algo:ns}. All Transformers have two-layer encoder and one-layer decoder with the feature dimension $d_k$=512, including Transformer~\cite{NIPS2017_3f5ee243}, Informer~\cite{haoyietal-informer-2021}, Reformer~\cite{kitaev2020reformer}, Autoformer~\cite{wu2021autoformer},
ETSformer~\cite{woo2022etsformer} and FEDformer~\cite{zhou2022fedformer}. Besides, we adopt embedding method and one-step generation strategy of Informer~\cite{haoyietal-informer-2021}. It is worth noting that for the row length of attention map differs from $S\times S$, where $S$ is the initial input sequence length, we omit the shifting de-stationary factor $\mathbf{\Delta}$ in Equation~\ref{equ:nsattns} (i.e., the Self-Attention layer of Transformer decoder, and the Self-Attention layer of the Informer encoder where the shape of attention map is changed over layers), since the performance of only use $\tau$ will not degenerate a lot as shown in Table~\ref{tab:full_abla}. For the cross attention, we first conduct the rescaling operation with de-stationary factors and then multiply by the corresponding mask. For Transformer variants, we conduct the rescaling operation on the pre-$\operatorname{Softmax}$ scores.

\begin{algorithm*}[htbp]
  \setstretch{1.5}
  \caption{Series Stationarization - Normalization.}\label{algo:ssn}
  \begin{algorithmic}[1]
    \Require  
     Input past time series $\mathbf{x}\in\mathbb{R}^{S\times C}$; Input Length $S$; Variables number $C$.
     \State $\mu_{\mathbf{x}} = \texttt{Mean}(\mathbf{x}, \texttt{dim=0})$ \Comment{$\mu_{\mathbf{x}}\in\mathbb{R}^{1\times C}$}
     \State $\sigma_{\mathbf{x}} = \texttt{Std}(\mathbf{x}, \texttt{dim=0})$ \Comment{$\sigma_{\mathbf{x}}\in\mathbb{R}^{1\times C}$}
     \State $\mathbf{x^\prime} = \texttt{Repeat}\big((1/\sigma_{\mathbf{x}}), \texttt{dim=0}\big) \odot \big(\mathbf{x}-\texttt{Repeat}(\mu_{\mathbf{x}}, \texttt{dim=0})\big)$ \Comment{Normalize to $\mathbf{x^\prime}\in\mathbb{R}^{S\times C}$}
         
    \State $\textbf{Return}\ \mathbf{x^\prime}, \mu_{\mathbf{x}}, \sigma_{\mathbf{x}}$ \Comment{Return stationarized input and original statistics}
  \end{algorithmic} 
\end{algorithm*} 

\begin{algorithm*}[htbp]
  \setstretch{1.5}
  \caption{Series Stationarization - De-normalization.}\label{algo:ssd}
  \begin{algorithmic}[1]
    \Require  
     Predicted time series $\mathbf{y^\prime}\in\mathbb{R}^{O\times C}$ by the base model; original statistics of input $\mu_{\mathbf{x}},\sigma_{\mathbf{x}}\in\mathbb{R}^{1\times C}$; Output Length $O$; Variables number $C$. 
     \State $\mathbf{y} = \texttt{Repeat}\big(\sigma_{\mathbf{x}}, \texttt{dim=0}\big) \odot \mathbf{y^\prime}+\texttt{Repeat}(\mu_{\mathbf{x}}, \texttt{dim=0})$
     \Comment{De-normalize to $\mathbf{y}\in\mathbb{R}^{O\times C}$}
    \State $\textbf{Return}\ \mathbf{y}$ \Comment{Return de-normalized output}
  \end{algorithmic} 
\end{algorithm*} 

\begin{algorithm*}[htbp]
  \setstretch{1.5}
  \caption{De-stationary Attention.}\label{algo:da}
  \begin{algorithmic}[1]
    \Require  
  Queries $\mathbf{Q}^{\prime}\in\mathbb{R}^{S\times d_{k}}$; 
  Keys $\mathbf{K}^{\prime}\in\mathbb{R}^{S\times d_{k}}$;
  Values $\mathbf{V}^{\prime}\in\mathbb{R}^{S\times d_{k}}$;
  De-stationary factors $\tau\in \mathbb{R}^{+}, \mathbf{\Delta} \in \mathbb{R}^{S\times 1}$;
  Input Length $S$; Feature dimension $d_{k}$.
    \State $\texttt{Output} = \texttt{Softmax}\left( 
        \big(
        \tau\ \mathbf{Q}^{\prime}\mathbf{K^{\prime}}^{\top}+\texttt{Repeat}(\mathbf{\Delta}, \texttt{dim=1})
        \big)  / \sqrt{d_k}
    \right) \mathbf{V}^{\prime} $
    \Comment{rescaling by $\tau$ and $\mathbf{\Delta}$}
    \State $\textbf{Return}\ \texttt{Output}$ \Comment{Return de-stationary attention output}
  \end{algorithmic} 
\end{algorithm*} 

\begin{algorithm*}[htbp]
  \setstretch{1.5}
  \caption{Non-stationary Transformers - Overall Architecture.}\label{algo:ns}
  \begin{algorithmic}[1]
  \Require  
  Input past time series $\mathbf{x}\in\mathbb{R}^{S\times C}$; Input Length $S$; Predict length $O$; Variables number $C$; Feature dimension $d_{k}$; Encoder layers number $N$; Decoder layers number $M$. Technically, we set $d_{k}$ as 512, $N$ as 2, $M$ as 1.
    \State $\mathbf{x^{\prime}}, \mu_{\mathbf{x}}, \sigma_{\mathbf{x}} = \texttt{Normaliztion}(\mathbf{x})$
    \Comment{$\mathbf{x}^\prime\in\mathbb{R}^{S\times C}, \mu_{\mathbf{x}}\in\mathbb{R}^{1\times C}, \sigma_{\mathbf{x}}\in\mathbb{R}^{1\times C}$}
    \State $\log \tau, \mathbf{\Delta} =\texttt{MLP}(\mathbf{x}, \mu_{\mathbf{x}}, \sigma_{\mathbf{x}})$
    \Comment{$\tau\in\mathbb{R}^{+}, \mathbf{\Delta}\in\mathbb{R}^{S\times 1}$}
    \State $\mathbf{x}_{\text{enc}}^{\prime}, \mathbf{x}_{\text{dec}}^{\prime} = \mathbf{x}^{\prime}, \texttt{ConCat}\big(\mathbf{x}_{\frac{S}{2}:S}^{\prime}, \texttt{Zeros}(O, C)\big)$
    \Comment{$\mathbf{x}^{\prime}_{\text{enc}}\in\mathbb{R}^{S\times C}, \mathbf{x}^{\prime}_{\text{dec}}\in\mathbb{R}^{(\frac{S}{2}+O) \times C}$}
    \State $\mathbf{x}^{0\prime}_{\text{enc}}=\texttt{Embed}(\mathbf{x}_{\text{enc}}^{\prime})$ \Comment{$\mathbf{x}^{0\prime}_{\text{enc}}\in\mathbb{R}^{S\times d_{k}}$}

    \State $\textbf{for}\ l\ \textbf{in}\ \{1,\cdots,N\}\textbf{:}$\Comment{Non-stationary Encoder}
    
    \State $\textbf{\textcolor{white}{for}}\ \mathbf{x}^{l-1\prime}_{\text{enc}} = \texttt{LayerNorm}\big(\mathbf{x}^{l-1\prime}_{\text{enc}} + \texttt{Attn}(\mathbf{x}_{\text{enc}}^{l-1\prime}, \tau, \mathbf{\Delta})\big)$
    \Comment{$\mathbf{x}^{l-1\prime}_{\text{enc}}\in\mathbb{R}^{S\times d_{k}}$}
    \State $\textbf{\textcolor{white}{for}}\ \mathbf{x}^{l\prime}_{\text{enc}} = \texttt{LayerNorm}\big(\mathbf{x}^{l-1\prime}_{\text{enc}} + \texttt{FFN}(\mathbf{x}_{\text{enc}}^{l-1\prime})\big)$
    \Comment{$\mathbf{x}^{l\prime}_{\text{enc}}\in\mathbb{R}^{S\times d_{k}}$}
    \State $\textbf{End for}$
    
    \State $\mathbf{x}^{0\prime}_{\text{dec}}=\texttt{Embed}(\mathbf{x}_{\text{dec}}^{\prime})$  \Comment{$\mathbf{x}^{0\prime}_{\text{dec}}\in\mathbb{R}^{(\frac{S}{2}+O)\times d_{k}}$}
    
    \State $\textbf{for}\ l\ \textbf{in}\ \{1,\cdots,M\}\textbf{:}$\Comment{Non-stationary Decoder}
    
    \State $\textbf{\textcolor{white}{for}}\ \mathbf{x}^{l-1\prime}_{\text{dec}} = \texttt{LayerNorm}\big(\mathbf{x}^{l-1\prime}_{\text{dec}} + \texttt{Attn}(\mathbf{x}_{\text{dec}}^{l-1\prime}, \tau, \mathbf{\Delta}=0)\big)$
    \Comment{$\mathbf{x}^{l-1\prime}_{\text{dec}}\in\mathbb{R}^{(\frac{S}{2}+O)\times d_{k}}$}
    
    \State $\textbf{\textcolor{white}{for}}\ \mathbf{x}^{l-1\prime}_{\text{dec}} = \texttt{LayerNorm}\big(\mathbf{x}^{l-1\prime}_{\text{dec}} + \texttt{Attn}(\mathbf{x}_{\text{dec}}^{l-1\prime}, \mathbf{x}_\text{enc}^{N\prime}, \tau, \mathbf{\Delta})\big)$
    \Comment{$\mathbf{x}^{l-1\prime}_{\text{dec}}\in\mathbb{R}^{(\frac{S}{2}+O)\times d_{k}}$}
    
    \State $\textbf{\textcolor{white}{for}}\ \mathbf{x}^{l\prime}_{\text{dec}} = \texttt{LayerNorm}\big(\mathbf{x}^{l-1\prime}_{\text{dec}} + \texttt{FFN}(\mathbf{x}_{\text{dec}}^{l-1\prime})\big)$
    \Comment{$\mathbf{x}^{l\prime}_{\text{dec}}\in\mathbb{R}^{(\frac{S}{2}+O)\times d_{k}}$}
    \State $\textbf{End for}$
    
    \State $\mathbf{y}^\prime = \texttt{MLP}(\mathbf{x}^{M\prime}_{\text{dec}})_{\frac{S}{2}:\frac{S}{2}+O}$
    \Comment{$\mathbf{y}^\prime\in\mathbb{R}^{O\times d_{k}}$}
    
    \State $\mathbf{y}= \texttt{De-normaliztion}(\mathbf{y^\prime}, \mu_{\mathbf{x}}, \sigma_{\mathbf{x}} )$
    \Comment{$\mathbf{y}\in\mathbb{R}^{O\times d_{k}}$}
    \State $\textbf{Return}\ \mathbf{y}$ \Comment{Return the prediction results}
  \end{algorithmic} 
\end{algorithm*}

\section{Broader Impact}\label{app:impact}

\subsection{Impact on Real-world Applications}
We focus on real-world time series forecasting, which is challenging for Transformers because of data non-stationarity. Our method goes beyond previous studies that only stationarize the time series. We fully utilize the predictive capability of attention mechanism that captures essential temporal dependencies associated with inherent non-stationarity. Our proposed method achieves state-of-the-art performance in five real-world applications, which makes it more promising for Transformers to tackle real-world forecasting applications, and helps our society make better decisions and prevent risks in advance for various fields. And our paper mainly focuses on scientific research and has no obvious negative social impact.

\subsection{Impact on Future Research}
In this paper, we analyze the generalization difficulty of Transformers in distribution-varying time series forecasting. We propose a general framework to fulfill the potential of Transforms constrained by data non-stationary. Our work introduces an essential and promising direction to improve forecasting performance: to increase the stationarity of time series towards better predictability and mitigate the over-stationarization problem for the predictive capability of deep models simultaneously. The remarkable generality and effectiveness of the proposed framework can be instructive for future research.

\section{Limitation}\label{app:limit}
Our De-stationary Attention is deduced by analyzing the vanilla Self-Attention, which may not be the optimal solution for advanced attention mechanisms. There also remains room for re-incorporating non-stationarity on other classical stationarization methods, like differencing and quantile. Besides, the proposed framework is currently limited to the Transformer-based models, while the over-stationarization problem can appear on any deep time forecasting models if using stationarization methods inappropriately. Therefore, a more model-agnostic solution for the over-stationarization problem will be our exploring direction.

\end{document}